\documentclass[lettersize,journal]{IEEEtran}

\usepackage{graphics} 
\usepackage{epsfig} 

\usepackage{times} 
\usepackage{mathrsfs}
\usepackage{colortbl}
\usepackage[table]{xcolor}

\usepackage{cite}
\usepackage{color}
\usepackage{multirow}
\usepackage{booktabs}
\usepackage{url}
\usepackage{amsmath,amssymb,amsfonts}
\usepackage{textcomp}
\usepackage{xcolor}
\usepackage{epstopdf}
\usepackage{subfigure}
\usepackage{stfloats}

\usepackage{pbox}
\usepackage{bm} 
\usepackage{multirow}
\usepackage{diagbox}
\usepackage{ulem}
\usepackage{booktabs}
\usepackage{makecell}

\usepackage{soul}

\usepackage[caption=false,font=normalsize,labelfont=sf,textfont=sf]{subfig}

\begin{document}

\title{\LARGE \bf
Active View Planning for Visual SLAM in Outdoor Environments Based on Continuous Information Modeling
}

\author{Zhihao~Wang, Haoyao~Chen*, \textit{Member, IEEE}, Shiwu~Zhang, \textit{Member, IEEE}, Yunjiang~Lou, \textit{Senior Member, IEEE}
  \thanks{This work was supported in part by the National Natural Science Foundation of China under Grant U1713206. (Corresponding author: Haoyao Chen.)}
  \thanks{Z.H. Wang, H.Y. Chen*, Y.J. Lou are with the School of Mechanical Engineering and Automation, Harbin Institute of Technology Shenzhen, P.R. China, e-mail: hychen5@hit.edu.cn.}
  \thanks{S.W. Zhang is with the Department of Precision Mechinery and Precision Instrumentation, University of Science and Technology of China.(swzhang@ustc.edu.cn)}
}

\markboth{IEEE/ASME TRANSACTIONS ON MECHATRONICS,~Vol.~xx, No.~xx, doi: 10.1109/TMECH.2023.3272910}%
{Shell \MakeLowercase{\textit{et al.}}: A Sample Article Using IEEEtran.cls for IEEE Journals}

\maketitle

\begin{abstract}

The visual simultaneous localization and mapping (vSLAM) is widely used in satellite-denied and open field environments for ground and surface robots. However, due to the frequent perception failures derived from featureless areas or the swing of robot view direction on rough terrains, the accuracy and robustness of vSLAM are still to be enhanced. The study develops a novel view planning approach of actively perceiving areas with maximal information to address the mentioned problem; a gimbal camera is used as the main sensor. Firstly, a map representation based on feature distribution-weighted Fisher information is proposed to completely and effectively represent environmental information richness. With the map representation, a continuous environmental information model is further established to convert the discrete information space into a continuous one for numerical optimization in real-time. Subsequently, the receding horizon optimization is utilized to obtain the optimal informative viewpoints with simultaneously considering the robotic perception, exploration and motion cost based on the continuous environmental model. Finally, several simulations and outdoor experiments are performed to verify the improvement of localization robustness and accuracy by the proposed approach. 
We release our implementation as an open-source\footnote{Code is available at https://github.com/HITSZ-NRSL/IGLOV.git} package for the community.

\end{abstract}

  
\vspace{-1em}
\section{Introduction}
Visual SLAM is widely used in outdoor and field environments for environmental monitoring, resource exploration and lakeshore inspection since the solution is of low cost and rich perceptual information\cite{9690581, 9830851}. 
However, the outdoor environments, like rough terrains and featureless water surface or ground, introduce difficulties for commonly used vSLAM algorithms\cite{murORB2} to achieve robust performance. 
For example, when the surface robot is floating along the lakeshore for inspection or cleaning, the featureless water surface will occupy large field-of-view (FOV);
when the ground robot is moving uphill, the perception system on the robot may tilt with the body to look towards the sky. 
These conditions easily lead to lost features and failure of localization. 
The inertial measurement unit (IMU) measurements and vision are fused to deal with feature loss in featureless and dynamic environments \cite{qin2018vins}. However, robot pose tracking often fails when working for a long time in the complex environments due to the loss of visual tracking\cite{wang2019robust}. 
Likewise, the large FOV cameras which can observe more features \cite{matsuki2018omnidirectional} are utilized for robust tracking in featureless scene; however the loss of angular resolution for higher FOVs is drastically amplified by the higher depth range in outdoor environments\cite{zhang2016benefit}, leading to the worse tracking performance than the perspective camera.

Animals with necks can turn their view flexibly to areas of interest. Similarly, actively controlling sensors like a camera to look at the places with rich features will benefit robotic navigation performance.
Besides, the robotic trajectory may be defined by the operator in remote monitoring and operating applications like lakeshore inspection. Thus, the view planning problem - where to look and how to look under the predefined trajectory - is essential to improve the estimation accuracy \cite{frintrop2008attentional}. 
To address the problem, this paper utilizes a gimbal camera which can turn the camera toward the best view with minimal estimation uncertainty. 
Note that the whole-body planning of camera view and robotic trajectory together is another research topic and is out of the research scope of this paper.

Generally, for the active visual SLAM, three challenging issues need to be considered \cite{chen2020active},
namely, how to represent the robotic state estimation uncertainty with respect to the camera view,
how to evaluate the performance of the candidate views to consider the trade-off of viewing new places (i.e., exploration) and reducing the estimation uncertainty by re-viewing known feature points (i.e., exploitation),
and how to select the next best view in all candidate views in real-time. 

The estimation uncertainty depends on many factors like texture, illumination \cite{wang2022automated}, and dynamic or static environmental objects. Various indicators like feature numbers \cite{deng2018feature}, and Fisher information \cite{khosoussi2019reliable, 2020arXiv200803324Z} are generally used for evaluating the estimation uncertainty. 
The feature numbers-based solution utilizes the numbers of feature points in the view to represent the estimation uncertainty. However, it is difficult to quantify the uncertainty of each feature point. 
The Fisher information-based solution is able to quantify the uncertainty meaningfully.
For the information computation, the existing approaches based on feature points\cite{khosoussi2019reliable} needs to calculate the Jacobian matrix of each feature, resulting in computation efficiency suffering. 
The localization information is summarized into voxels by using Fisher information field to accelerate the computation for online planning, but the building time of the field is still higher due to the iteration of each feature\mbox{\cite{2020arXiv200803324Z}}.
Voxelization\mbox{\cite{hornung2013octomap}} can be used to accelerate the map building process by downsampling the feature points, but the voxelization concentrates on the occupancy of each voxel and neglects point distribution in the voxel. This neglect results in the ambiguity of the estimation uncertainty representation when calculating Fisher information with voxels. 
The ambiguity is described in Section \mbox{\ref{section3}} in detail.
Therefore, the accurate and efficient representation of environmental estimation uncertainty for online mapping and planning remains challenging.

Besides the information representation, the evaluation of candidate views for robotic exploration/exploitation is also an essential problem for active SLAM.
For exploration, many researchers utilize lasers or cameras to detect geometric frontiers or calculate information gain to plan the sensor movement \cite{jadidi2018gaussian}; the purpose is to completely explore the unknown environments as soon as possible. 
For exploitation, some approaches plan a feature-rich trajectory to minimize the state estimation uncertainty with a camera fixed on a quad-rotor \cite{zhang2018perception} or a gimbal camera on a mobile robot \cite{strader2020perception}.
However, due to the greedy consideration of the minimal localization uncertainty, 
the existing methods \cite{zhang2018perception, strader2020perception} suffer from continuously revisiting known areas without exploring unknown areas. This leads to the degeneration \cite{strader2020perception} or local minimum problem \cite{chen2020active} of planning, especially for the view planning in unknown environments. 
Some approaches have been developed by utilizing a mode switching mechanism \cite{frintrop2008attentional, kim2015active, chen2020active} to address the exploration-exploitation dilemma.
A solution presented in \cite{frintrop2008attentional} involved switching the exploitation mode to the exploration mode when sufficient landmarks are successfully detected.
Different weights were assigned to the modes in \cite{kim2015active, chen2020active}, and balancing the SLAM uncertainty reduction and area coverage task performed well. 
{By contrast, this study utilizes motion consistency\mbox{\cite{zhou2021fuel}} as an exploration indicator to deal with the degeneration problem of view planning. The motion consistency produces attractive force to keep the next best view being consistent with the motion direction.}
Subsequently, the evaluation of candidate views considers both the exploitation of information and the exploration indicator in an objective function; this makes it possible to solve the exploration-exploitation dilemma with continuous planning. 

Based on the evaluation of candidate views, sample-based methods, e.g., RRT* {\cite{strader2020perception}}, or Dynamic movement primitive{\cite{dharmadhikari2020motion}} were developed to select the best view with maximal utility. The methods suffer from discontinuous motion due to the discrete sampling.
Some researchers\mbox{\cite{zhu2021online, indelman2015planning}} realized continuous-space planning by maximizing the utility of candidate states. 
However, the utility function of informative path planning is always high-dimensional, nonlinear, and non-convex, which makes solving the optimization problem of maximizing the utility difficult. 
The evolutionary algorithm is used to solve the complex optimization problem\mbox{\cite{zhu2021online}}, but it is time-consuming. 
The gradient descent is an efficient optimization technology but hard to derive an analytic expression of the complex utility function's gradient\mbox{\cite{indelman2015planning}}.

The paper aims to develop a novel approach by actively and smoothly controlling a gimbal camera equipped on the robot to realize robust and accurate SLAM in unknown outdoor environments. 
The three challenging issues mentioned above are solved efficiently in the proposed approach. The contributions of the paper are three-fold. 

{First, a novel map representation based on feature distribution-weighted Fisher information is proposed to store the localization uncertainty of environments. This new information map overcomes the ambiguity of the localization uncertainty representation of the traditional voxelization method. Our method makes the environmental information representation more accurate and efficient for active perception and further helps realize online information mapping and motion planning. }

{Second, a continuous information modeling method is proposed to map the environmental information, like localization uncertainty around the robot, into multiple polynomial functions. The polynomial functions provide analytic derivatives for the environmental information with respect to the action space, and therefore makes the informative planning problem be solved efficiently by numerical optimization with lightweight time-consuming. }

{Third, an information gradient-based local view (IGLOV) planner is proposed to plan the optimal camera views in real-time for obtaining maximal environmental information. The planner realizes active view planning by considering the estimation uncertainty, exploration for avoiding degeneration and motion smoothness constraints simultaneously. The experiments illustrate that our approach outperforms the state of the art.}

\section{System Overview}\label{section2}

\begin{figure}[htbp]
  \setlength{\abovecaptionskip}{0.cm} 
  \centering
  \centerline{\includegraphics[width=3.0in]{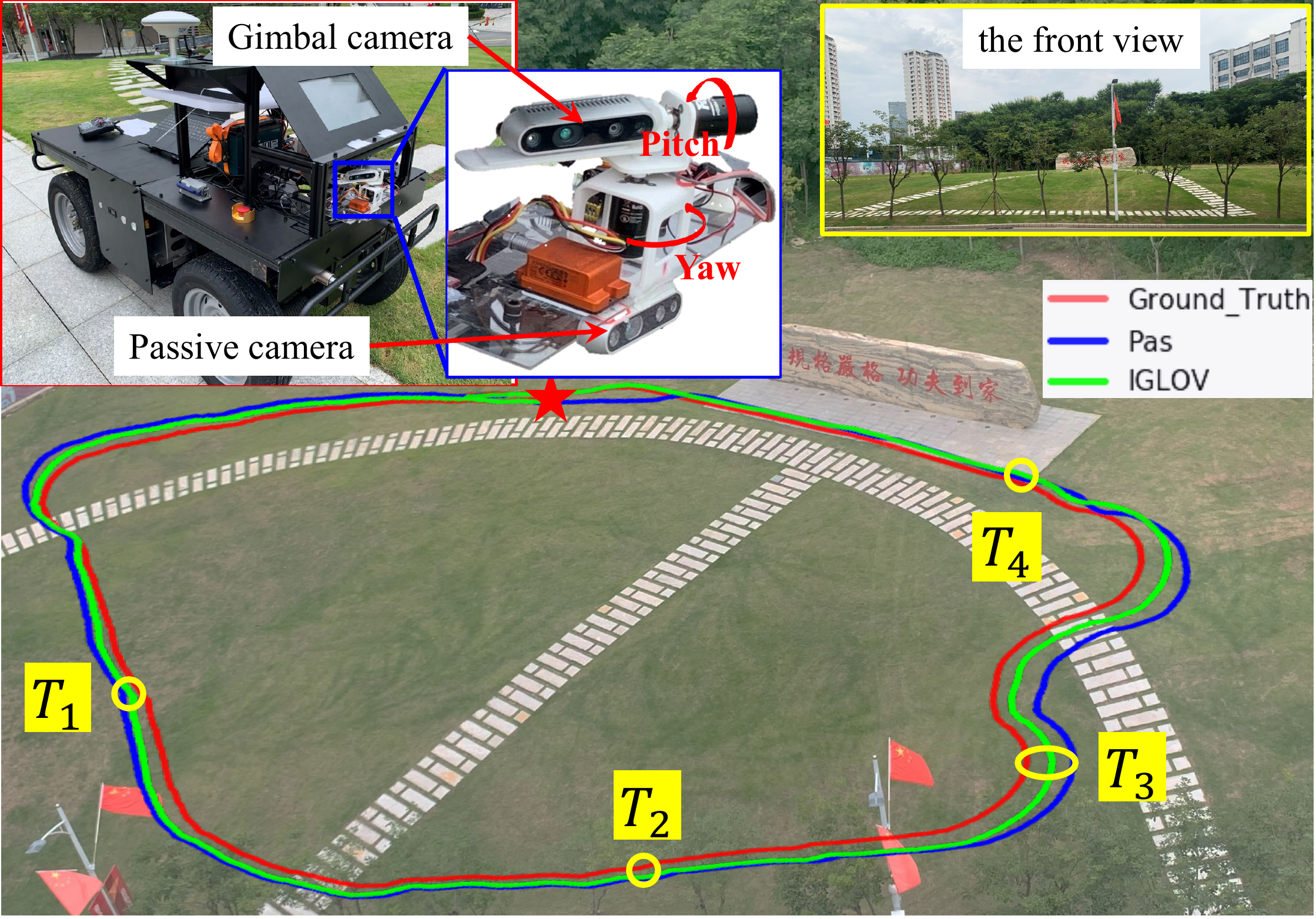}}
  \caption{ Illustration of active view planning of a gimbal camera for visual SLAM in an outdoor environment.    }
  \label{fig:Problem Formulation}
  \vspace{-0.2cm}  
\end{figure}

This work considers a ground or surface robot traveling along predefined trajectories in unknown outdoor environments, as illustrated in Fig.~\ref{fig:Problem Formulation}. A camera is equipped on the robot through a two-axis gimbal. To guarantee the robustness of trajectory tracking, a robust visual SLAM is required by automatically changing the camera perception direction to achieve stable state estimation. 
Therefore, the problem to be addressed in this work is formulated as
\begin{equation}\label{eq:problem formulation}
  \setlength{\abovedisplayskip}{3pt}
  \setlength{\belowdisplayskip}{3pt}  
  \begin{aligned}
    \bm{u}^* := &\underset{\bm{u}}{\arg \max }  \  f\left(\bm{{\xi}}^\mathrm{wc}, \mathcal{M}, \bm{u}\right) \\
  \text{s.t.   } &{h}(\bm{{\xi}}^\mathrm{wc}, \bm{u}) \leq 0
  \end{aligned}
\end{equation}
where $f(\cdot)$ denotes an objective function to be designed that quantifies the estimation accuracy of the camera pose $\bm{{\xi}^}\mathrm{wc}$; $\mathcal{M}$ denotes the estimated map; {$\bm{u}$ including yaw and pitch control commands denotes the control vector of the gimbal to be optimized;} $\bm{u}^*$ is the optimal control vector of $\bm{u}$; $h(\cdot)$ represents the constraints for the gimbal camera; $\bm{{\xi}}^\mathrm{wc} = [\bm{{\chi}}^\mathrm{wc}, \bm{{\phi}}^\mathrm{wc}]^{\mathrm{T}} \in \mathbb{R}^{6}$ denotes the camera pose with respect to the world frame, where $\bm{{\chi}}^\mathrm{wc} \in \mathbb{R}^{3}$ and $\bm{{\phi}}^\mathrm{wc} \in \mathbb{R}^{3}$ denote the translation and rotation, respectively.

\begin{figure}[htbp]
  \setlength{\abovecaptionskip}{0.cm} 
	\centerline{\includegraphics[width=3.5in]{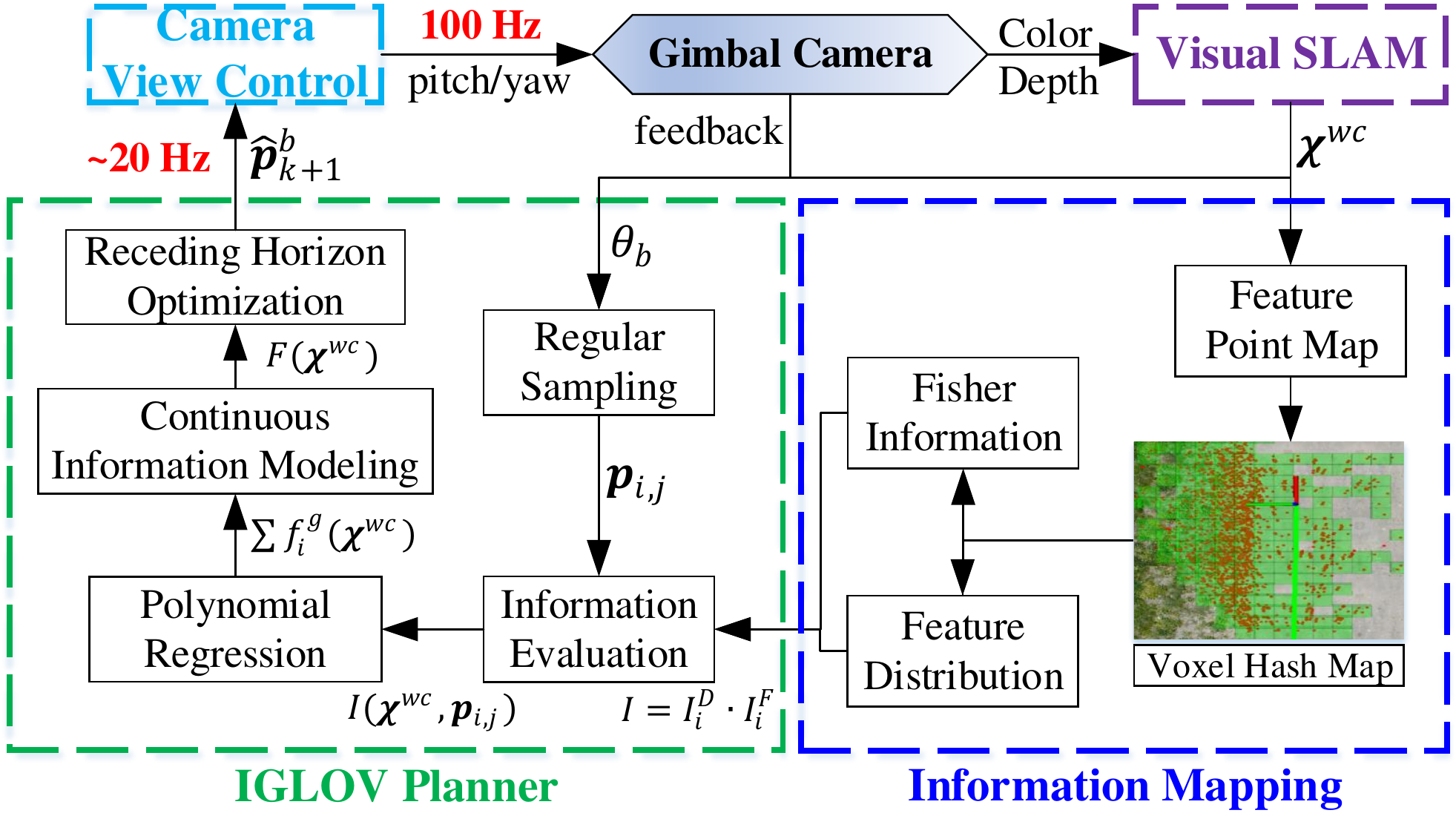}}
  \caption{Framework of the proposed active visual SLAM approach. }
	\label{fig:Framework of the active SLAM system}
  \vspace{-0.2cm}  
\end{figure}

Figure \ref{fig:Framework of the active SLAM system} illustrates a framework of active visual SLAM by integrating the proposed solutions of information mapping, continuous environmental modeling and camera view planning. 
The camera poses together with the feature point map are obtained using the common visual SLAM. 
The information mapping process lies in evaluating the information richness of the environment;
to achieve the objective, a mapping method that includes both Fisher information and feature distribution information is proposed. We use the Fisher information to evaluate the estimation uncertainty stored in the voxel like\cite{2020arXiv200803324Z}, and use feature distribution information to evaluate feature number and distribution in the voxel.  
Based on the above information mapping, the camera view planning algorithm is then developed to obtain the camera's best view direction; the direction is represented by the means of a view-landing-point.
The algorithm consists of four steps, including the generation of sample points, the evaluation of information gain, polynomial regression, and receding horizon optimization. 
Finally, the best view-landing-point obtained from the algorithm is sent to the bottom tracking controller of the gimbal.

\section{Environmental Information Mapping}\label{section3}

The information richness of the surrounding environment should be evaluated to guide the camera view planning. 
The voxelization is used to efficiently represent the information richness of the environment by downsampling the feature point cloud.
{However, the representation with voxelization raises a new problem in active perception. 
The information calculated from the voxel map represents the localization uncertainty of each voxel but neglects the effect of the feature number in the voxel and the feature distribution around the voxel on the localization uncertainty. Especially for the feature-based vSLAM, the feature number and their distribution uniformity directly affect the tracking accuracy\mbox{\cite{qin2018vins}}.
Specifically, the neglect of feature distribution results in feature-dense areas that have the same information as the feature-sparse areas; however, feature-dense areas actually contain more features for tracking than feature-sparse areas. This problem is called as the ambiguity in the representation of estimation uncertainty brought by voxelization.}
{To deal with the ambiguity problem, the feature distribution information, including the feature number in the voxel and the feature distribution around the voxel, is integrated with the Fisher information of a voxel. 
The integration of the two kinds of information is realized by a new map representation method to evaluate the information richness of the environment.}

\vspace{-1.4em}
\subsection{Calculation of Fisher information}
The Fisher information matrix (FIM) indicates the $Cram\acute{e}r-Rao$ lower bound, the smallest covariance of an unbiased estimator \cite{barfoot2017state}. Therefore, the Fisher information matrix is usually used to represent the estimation uncertainty in many robotic applications like feature selection \cite{chen2021anchor}.

{As our approach focuses on the view planning based on the known feature point maps obtained from SLAM, the uncertainty of camera pose estimation is evaluated from the perspective of observation uncertainty to decide the next best view. 
The observation uncertainty is evaluated by the Fisher information that carry about estimating the camera pose $\bm{{\xi}}^\mathrm{wc}$.
The observation $\bm{z}_{i}$ of $\bm{p}_{i}^\mathrm{w}$ at the camera pose $\bm{{\xi}}^{\mathrm{wc}}$ is modeled as}
\begin{equation}\label{eq:specific observation function}
  \setlength{\abovedisplayskip}{3pt}
  \setlength{\belowdisplayskip}{3pt}  
  \bm{z}_{i} = {\mathrm {g}}(\bm{{\xi}}^{\mathrm{wc}}, \bm{p}_{i}^\mathrm{w}) + \bm{\omega} 
\end{equation}
{where \mbox{$\mathrm{g}(\cdot)$} is derived from the camera's measurement model; here the bearing vector model is implemented as the measurement model\mbox{\cite{zhang2019beyond}} and is defined as} 
\begin{equation}
  \mathrm{g}\left(\boldsymbol{\xi}^{\mathrm{wc}}, \boldsymbol{p}_i^{\mathrm{w}}\right)={\mathbf{p}_i^c}/{\left\|\mathbf{p}_i^c\right\|_2}, \mathbf{p}_i^c = {(\mathrm{exp}({\boldsymbol{\xi}^{\mathrm{wc}}}^{\wedge})})^{\mathrm{T}} \mathbf{p}_i^{\mathrm{w}}.
\end{equation}
{where $\bm{p}_{i}^\mathrm{w}$ and $\bm{p}_{i}^\mathrm{c}$ denotes the center position of the $i$th voxel $V_{i}$ in the world coordinate frame and the camera coordinate frame, respectively.} $i={0,1,...,M}$ and $M$ is the number of occupied voxels of the voxel map $\mathcal{M}$; {$\bm{\omega}$} denotes white noise with covariance $\bf{Q}$.
{Then, the Fisher information matrix $\bm{I}_{i} \in \mathbb{R}^{6\times6}$ evaluates the estimation uncertainty derived from the observation of $\bm{p}_{i}^\mathrm{w}$ at $\bm{{\xi}}^{\mathrm{wc}}$, and is defined as:}
\begin{equation}\label{eq:fisher information jacobian}
  \setlength{\abovedisplayskip}{3pt}
  \setlength{\belowdisplayskip}{3pt}  
  \begin{aligned}
  \bm{I}_{i}=\bm{J}_{\mathrm{g}}^{\mathrm{T}}\bm{Q}^{-1}\bm{J}_{\mathrm{g}}
  \end{aligned}
\end{equation}
where $\bm{J}_{\mathrm{g}}=\partial {\mathrm{g}}/{\partial \bm{{\xi}}^\mathrm{wc}}$ denotes the Jacobian of the observation function $\mathrm{g}(\cdot)$ with respect to $\bm{{\xi}}^\mathrm{wc}$. {For detailed deduction, please refer to\mbox{\cite{bishop2010optimality}}}.
Because $\{\bm{I}_{0}, ... \bm{I}_{M}\}$ is a series of matrices, the memory usage increases rapidly with the environmental exploring. A common way to evaluate voxels in terms of estimation accuracy is based on the theory of optimal experimental design (TOED) \cite{pukelsheim2006optimal}. TOED utilizes the T-opt optimality criterion, i.e., the trace of the FIM, to convert the matrix to a scalar metric for reducing memory usage.
Moreover, it has been proved that the Fisher information matrix without the visibility constraint is rotation-invariant \cite{zhang2019beyond}; that means the FIM only relates to the camera position $\bm{{\chi}}^\mathrm{wc}$, and not concerned with the camera rotation $\bm{{\phi}}^\mathrm{wc}$. 
{Therefore, we finally define the Fisher information metric $\mathnormal{I}_{i}^{F}$ as}
\begin{equation}\label{eq:trace of fisher information without rotation}
  \setlength{\abovedisplayskip}{3pt}
  \setlength{\belowdisplayskip}{3pt}  
  \mathnormal{I}_{i}^{F}\left(\bm{{\chi}}^\mathrm{wc},  \bm{p}_{i}^\mathrm{w}\right)=\operatorname{\textrm{trace}}\left(\bm{I}_{i}\right)
\end{equation}
{where $\mathnormal{I}_{i}^{F}\left(\bm{{\chi}}^\mathrm{wc},  \bm{p}_{i}^\mathrm{w}\right)$ represents the estimation uncertainty of $\bm{{\chi}}^\mathrm{wc}$ when observing $V_{i}$ at $\bm{{\chi}}^\mathrm{wc}$.}

\vspace{-1.4em}
\subsection{Calculation of distribution-weighted Fisher information}

The uniformity of feature distribution affects the tracking accuracy in the feature-based SLAMs.
The statistics of features in the neighbor voxel set $S_{ne}$, which consisted of the 27 neighbor voxels around $V_{i}$, are used to formulate the uniformity of feature distribution.
Specifically, the mean and the standard deviation of the feature number in $S_{ne}$ are defined as
\begin{equation}\label{eq:mean of feature number}
  \setlength{\abovedisplayskip}{3pt}
  \setlength{\belowdisplayskip}{3pt}  
  \mu_{i} = \frac{1}{N_{ne}} \sum_{k=1}^{N_{ne}} \mathnormal{N}_{k}^{D}
\end{equation}
\begin{equation}\label{eq:deviation of feature number}
  \setlength{\abovedisplayskip}{3pt}
  \setlength{\belowdisplayskip}{3pt}  
  \sigma_{i}=\sqrt{\frac{\sum_{k=1}^{N_{ne}}\left(\mathnormal{N}_{k}^{D}-\mu_{i}\right)^{2}}{N_{ne}}}
\end{equation}
where $N_{ne}$ is the element number in $S_{ne}$;
$\mathnormal{N}_{k}^{D}$ denotes the feature number within the $k$-th neighbor voxel $V_{k}$ in $S_{ne}$.
The uniformity $\mathnormal{I}_{i}^{D}$ of feature distribution around $V_{i}$ is defined as
\begin{equation}\label{eq:feature distribution}
  \setlength{\abovedisplayskip}{3pt}
  \setlength{\belowdisplayskip}{3pt}  
  \mathnormal{I}_{i}^{D} = \mu_{i} \cdot (1+e^{-\sigma_{i}}).
\end{equation}

Because the calculation of Fisher information is based on the discrete voxels, which neglects the feature number and distribution around the voxel, we complement the Fisher information with the feature distribution of each voxel for accurate representation. 
$\mathnormal{I}_{i}^{F}$ concentrates on the estimation uncertainty of the camera pose in a voxel. While $\mathnormal{I}_{i}^{D}$ focuses on the local feature distribution around the voxel, which explicitly represents the localization uncertainty from the perspective of feature matching and tracking.
Then the information in \eqref{eq:trace of fisher information without rotation} and \eqref{eq:feature distribution} are fused as
\begin{equation}\label{eq:total_information}
  \setlength{\abovedisplayskip}{3pt}
  \setlength{\belowdisplayskip}{3pt}  
  \mathnormal{I}\left(\bm{{\chi}}^\mathrm{wc}, \bm{p}_{i}^\mathrm{w}\right)=\mathnormal{I}_{i}^{D} \cdot \mathnormal{I}_{i}^{F}\left(\bm{{\chi}}^\mathrm{wc}, \bm{p}_{i}^\mathrm{w}\right)  
\end{equation}
{where $\mathnormal{I}\left(\bm{{\chi}}^\mathrm{wc}, \bm{p}_{i}^\mathrm{w}\right)$ denotes the distribution-weighted Fisher information of $\bm{{\chi}}^\mathrm{wc}$ when observing $V_{i}$ at $\bm{{\chi}}^\mathrm{wc}$. }

{Based on \mbox{\eqref{eq:total_information}}, the distribution-weighted Fisher information by considering both the Fisher information and the feature distribution is able to correctly quantify the estimation uncertainty of camera poses. It provides the essential information metric for the camera view planning module in the following section.}

\section{Camera View Planning}\label{section4}
Literature\cite{kim2015active} has shown that the known areas of the map have more known features for tracking and contribute to low uncertainty, whereas unknown areas generally have fewer features and lead to high uncertainty. 
Therefore, the maximum information-based solution allows the robot to only revisit the known areas for robust localization. This condition may lead to the degeneration of navigation, especially for the active camera view planning in unknown environments.
To address the problem,  
we develop a novel information gradient-based local view (IGLOV) planner to actively minimize localization uncertainty while considering the degeneration simultaneously. 
The IGLOV planner contains four main parts, i.e., generating sample points, evaluating information gain, conducting polynomial regression, and receding horizon optimization. 

As the gimbal camera has only two degrees of freedom, solving the inverse kinematic is convenient. Furthermore, its view planning in task space can benefit the motion prediction and handling of environmental perception. 
Therefore, the planner optimizes the view-landing-points in task space; the view-landing-point means the intersection of the terrain surface and the camera optical axis. 
Followed by the gimbal's inverse kinematics, the view-landing-point will be transformed into desired gimbal rotation angles. 
Besides, "point" is used to denote "view-landing-point" for simplification in the rest of the paper.

\begin{figure}[htbp]
  \setlength{\abovecaptionskip}{0.cm} 
	\centerline{\includegraphics[width=2.6in]{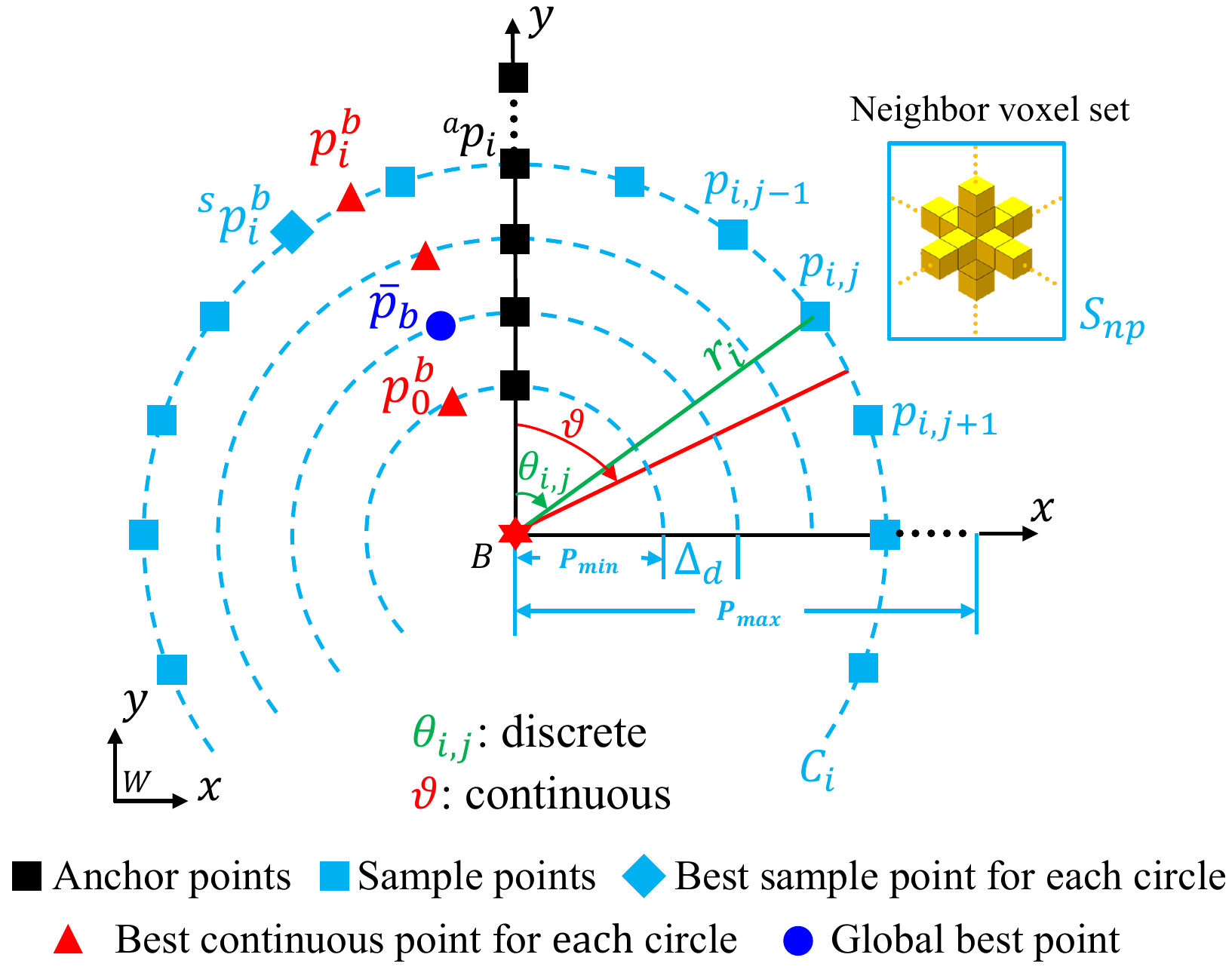}}
  \caption{View-landing-point sampling process in the IGLOV planner. }
	\label{fig:planner}
  \vspace{-0.5cm}  
\end{figure}

\vspace{-1em}
\subsection{Generation of sample points}
\label{Generation of sample points}
As shown in Fig.\ref{fig:planner}, the red star denotes the current pose of the robot, 
and frames W and B represent the world and the robot base coordinate frames, respectively.
The y-axis of $X_B-O_B-Y_B$ is along the robot's moving direction. 
The anchor points are firstly generated along the y-axis.
Moreover, each anchor point $^{a}\bm{p}_{i}$ corresponds to a sample circle $C_i$ for generating the sample points. 
Sample points are uniformly generated from $^{a}\bm{p}_{i}$ to both sides along the sample circle $C_i$ with radius $r_i$ and angle interval ${\Delta}_{\theta}$.
The $j$th sample point on $i$th sampling circle $C_i$ is denoted as $\bm{p}_{i,j}$, and its coordinates w.r.t. the world frame is given as
\begin{equation}\label{eq:generate sample}
  \setlength{\abovedisplayskip}{3pt}
  \setlength{\belowdisplayskip}{3pt}  
  \bm{p}_{i,j} = \bm{{\chi}}^\mathrm{wc} + [r_i*sin({\theta}_{i,j} + {\theta}_b), r_i*cos({\theta}_{i,j} + {\theta}_b), 0]^{\mathrm{T}}
\end{equation}
where $r_i = P_{min} + i * {\Delta}_{d}$, $i = 0, 1, ... , N_{AP}$, $P_{min}$ is the minimal range for the anchor points, ${\Delta}_{d}$ is the distance interval between two neighboring anchor points along the radial direction, $N_{AP}$ is the number of anchor points;
${\theta}_{i,j} = k * {\Delta}_{\theta}$, $k = -N_{SP}/2, ..., -1, 0, 1, ... , N_{SP}/2$, $j = 0, 1, ... , N_{SP}$, $N_{SP}$ is the number of sample points along $C_i$, 
$ {\theta}_{i,j} $ is the angle between $\bm{p}_{i,j}$ and $^{a}\bm{p}_{i}$, and ${\Delta}_{\theta}$ is the angle interval between two neighboring sample points (e.g., $\bm{p}_{i,{j-1}}$ and $\bm{p}_{i,j}$);
${\theta}_b$ is the yaw angle of the robot base calculated from $\bm{{\phi}}^{wb}$, and $\bm{{\phi}}^{wb}$ is the orientation of the robot base.

\vspace{-1em}
\subsection{Evaluation of information gain}
\label{Establishment of information gain function}
To evaluate the information gain of the sample points, a function is constructed with considering both the localization uncertainty and degeneration. The localization uncertainty is quantified by the information defined in \eqref{eq:total_information}. 
Further, the consistency between the view direction of the gimbal camera and the motion direction of the robot base is considered to deal with the degeneration. 
Thus, the information gain function of the sample point $\bm{p}_{i,j}$ is defined as
\begin{equation}\label{eq:cost sample}
  \setlength{\abovedisplayskip}{3pt}
  \setlength{\belowdisplayskip}{3pt}  
  g_{i,j} = I(\bm{{\chi}}^\mathrm{wc}, \bm{p}_{i,j}) - {\lambda}_d \cdot | {\theta}_{i,j} |
\end{equation}
where $I(\bm{{\chi}}^\mathrm{wc}, \bm{p}_{i,j})$ denotes the information of the voxel $\bm{p}_{i,j}$ calculated by \eqref{eq:total_information}; $|{\theta}_{i,j}|$ denotes the absolute value of ${\theta}_{i,j}$; 
{${\lambda}_d := I(\bm{{\chi}}^\mathrm{wc}, \bm{p}_{i,j}) / \pi$ is a dynamic weight coefficient for balancing the two terms into the same magnitude}.
The term $|{\theta}_{i,j}|$ represents the consistency between the view direction corresponding to the sample point $\bm{p}_{i,j}$ and the robotic motion direction, as demonstrated in Fig.\ref{fig:planner}. 
Thus, it penalizes the sample points deviating from the motion direction.

Equation \eqref{eq:cost sample} calculates the information gain of the single view-landing-point $\bm{p}_{i,j}$; however, it raises the problem that the information of a single point cannot wholly represent the information within the camera's FOV.
To this end, we consider the information around the sample point $\bm{p}_{i,j}$ by involving the neighboring voxel set $S_{np}$ within a predefined distance threshold $d_{n}$. 
Through the addition of $S_{np}$ into the information gain function, the planner takes the 3D environmental information into consideration even though with a 2D sampling method.
Finally, the information gain function \eqref{eq:cost sample} is rewritten as
\begin{equation}\label{eq:cost candidate extend}
  \setlength{\abovedisplayskip}{3pt}
  \setlength{\belowdisplayskip}{3pt}  
  \begin{aligned}
    & g_{i,j} = \sum_{\bm{p}_{np} \in S_{np}} I(\bm{{\chi}}^\mathrm{wc}, \bm{p}_{np} ) - {\lambda}_d \cdot | {\theta}_{i,j} |
  \end{aligned}
\end{equation}
{where $\bm{p}_{np}$ denotes an element in the set $S_{np}$.}

\vspace{-1em}
\subsection{Polynomial regression for continuous information gain}
The camera view planning is modeled as an optimization problem by evaluating the points in the task space. 
However, the information gain function defined in \eqref{eq:cost candidate extend} is discrete. 
For the convenience of numerical optimization, the function to be optimized must be continuous and differentiable. 
To address the problem, we apply the methodology of polynomial regression to approximate a continuous and differentiable function about the environmental information. 
Compared with Gaussian process regression and other surface fitting methods, the polynomial regression for each curve is efficient, with the mathematics expression being differentiable.
After the polynomial regression, the best point of the local environment can be obtained by searching the point with the maximal information gain in the multiple polynomial curves.
Besides, the optimal result based on multiple curves is approximately equivalent to the result on a surface when the distance ${\Delta}_{d}$ between two adjacent curves is small enough.

For each sample circle $C_i$, a polynomial function $f_i^g$ is obtained by fitting the information gain of all the discrete sample points $\bm{{p}}_{i,j}$ along $C_i$. 
The position of $\bm{p}_{i,j}$ is determined by $\theta_{i,j}$ from \eqref{eq:generate sample}, and thus the gain $g_{i,j}$ relates the angle $\theta_{i,j}$ according to \eqref{eq:cost candidate extend}. Therefore, we define the continuous information gain function $f_i^g$ as
\begin{equation}\label{eq:polynomial function}
  \setlength{\abovedisplayskip}{3pt}
  \setlength{\belowdisplayskip}{3pt}  
  \begin{aligned}
    f_i^g(\vartheta) & = {a}_{i,0} + {a}_{i,1} \cdot \vartheta + {a}_{i,2} \cdot {\vartheta}^2 + ... + {a}_{i,n} \cdot {\vartheta}^n 
  \end{aligned}
\end{equation}
where ${\vartheta}$, the domain of function $f_i^g$, denotes the continuous scanning angle variable along $C_i$, as shown in Fig.\ref{fig:planner}; 
$\bm{A}_i :=  ({a}_{i,0}, {a}_{i,1}, \dots, {a}_{i,n})^T \in \mathbb{R}^{n+1}$ denotes the stacked weight parameters to be calculated, 
and $n$ denotes the degree of the polynomial function.

The information gain values of all the sample points along $C_i$ should fit the function \eqref{eq:polynomial function}.
By stacking the relative sample points, we have
\begin{equation}
  \setlength{\abovedisplayskip}{3pt}
  \setlength{\belowdisplayskip}{3pt}  
  \bm{G}_i =  \bm{\varTheta}_i \bm{A}_i,
\end{equation}
where $\bm{\varTheta}_i$ denotes the Vandermonde matrix of ${\theta}_{i,j}, j \in (0,1,2, \dots, N_{SP})$, given as
\begin{equation}\label{eq:matrix of theta}
  \setlength{\abovedisplayskip}{3pt}
  \setlength{\belowdisplayskip}{3pt}  
  \begin{aligned}
    \bm{\varTheta}_i: = \left[\begin{array}{ccccc}1 &  {\theta}_{i,0} &  {\theta}_{i,0}^{2} & \cdots &  {\theta}_{i,0}^{n} \\  1 &  {\theta}_{i,1} &  {\theta}_{i,1}^{2} & \cdots &  {\theta}_{i,1}^{n} \\  1 &  {\theta}_{i,2} &  {\theta}_{i,2}^{2} & \cdots &  {\theta}_{i,2}^{n} \\ \vdots & \vdots & \vdots & \ddots & \cdots \\  1 &  {\theta}_{i,N_{SP}} &  {\theta}_{i,N_{SP}}^{2} & \cdots &  {\theta}_{i,N_{SP}}^{n}\end{array}\right];
  \end{aligned}
\end{equation}
$\bm{G}_i := (g_{i,0}, g_{i,1}, \dots, g_{i,N_{SP}})^T$, which is calculated by \eqref{eq:cost candidate extend}.

\begin{figure}[htbp]
  \setlength{\abovecaptionskip}{0.cm} 
  \vspace{-0.3cm}  
  \centering
  \subfigure[Sampling in Cartesian frame]{\includegraphics[width=1.65in]{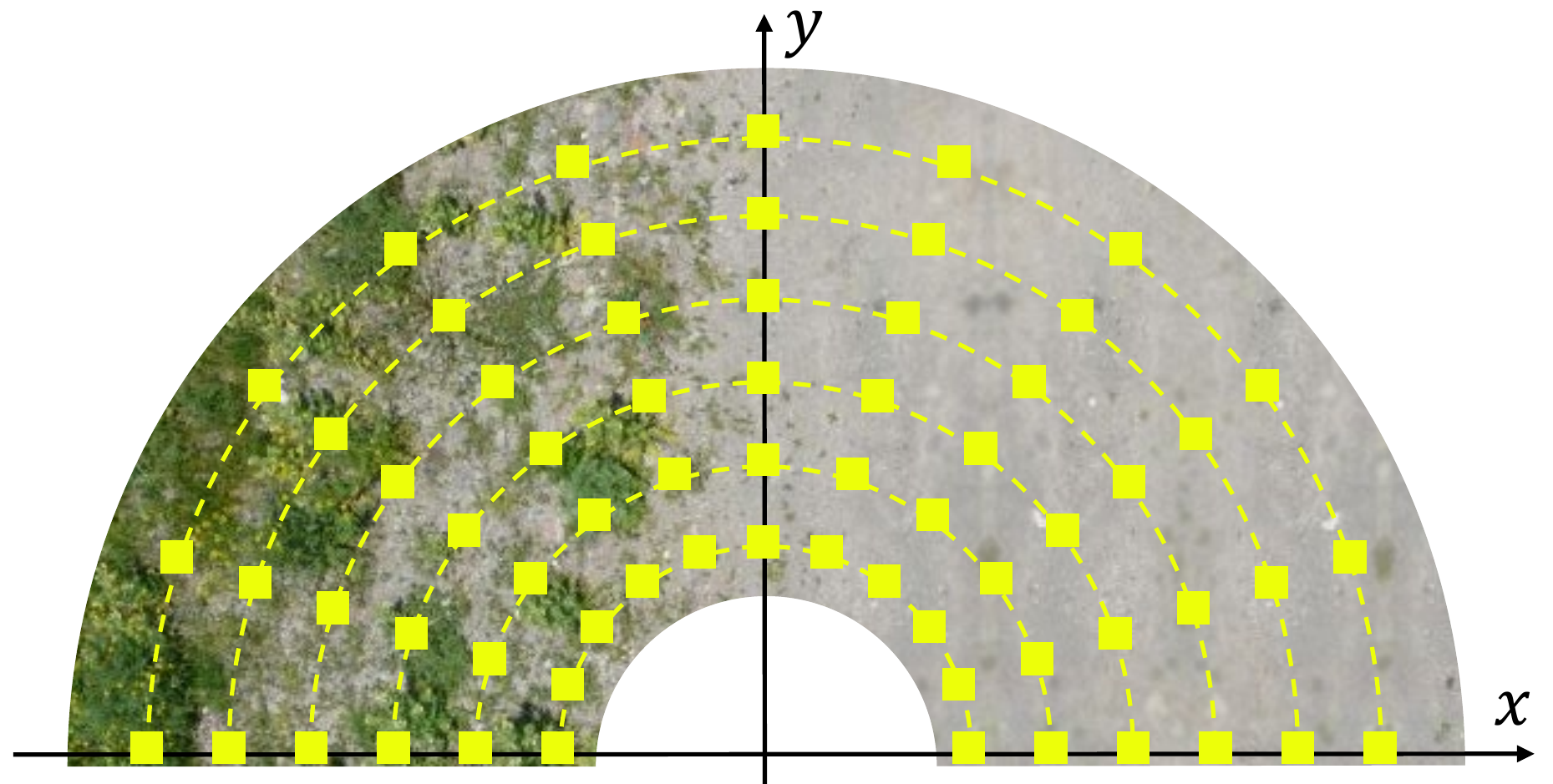}}
  \subfigure[Sampling in polar frame]{\includegraphics[width=1.65in]{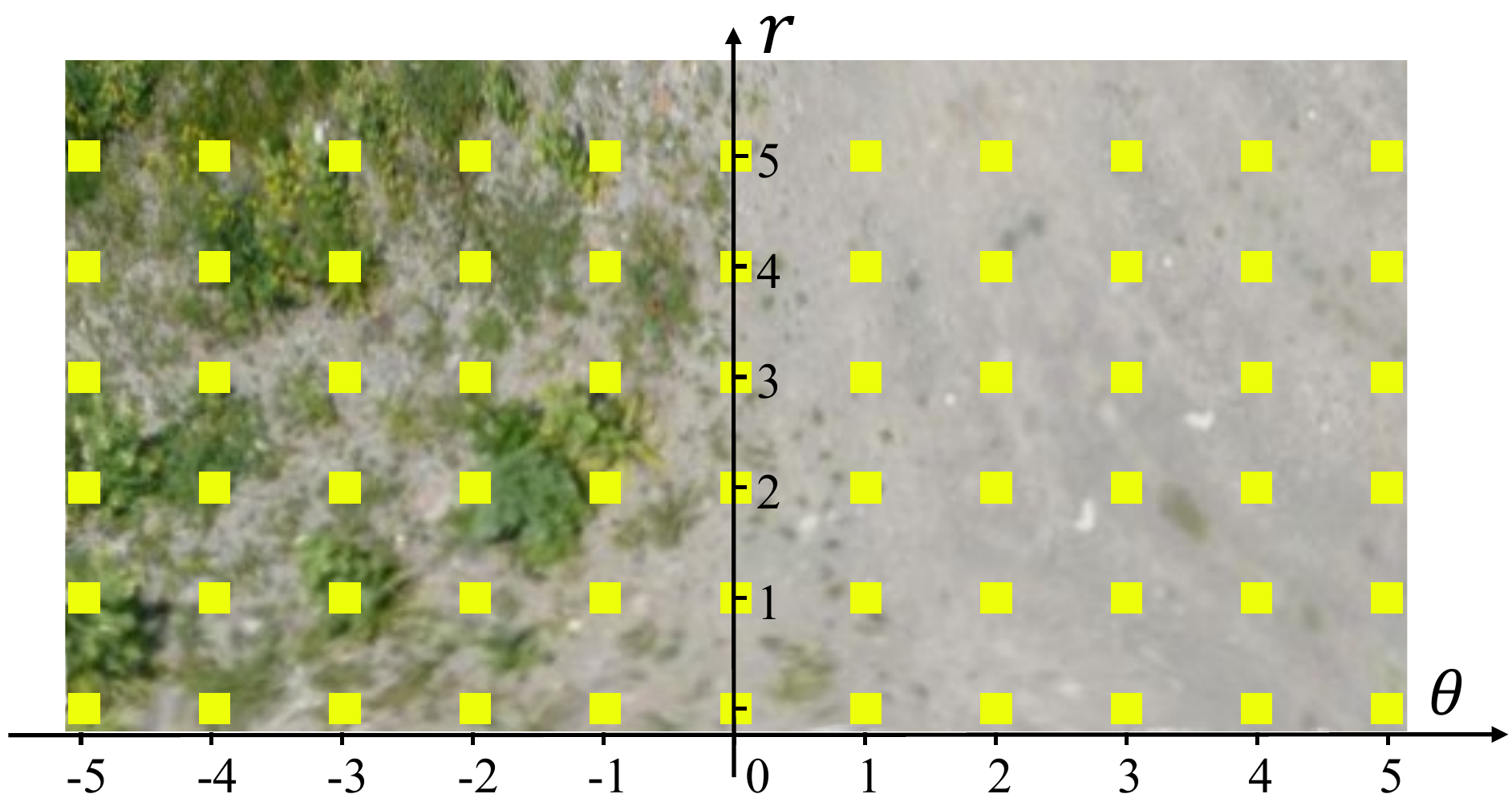}}
  \\
  \centering
  \subfigure[{Gain of sample points}]{\includegraphics[width=1.65in]{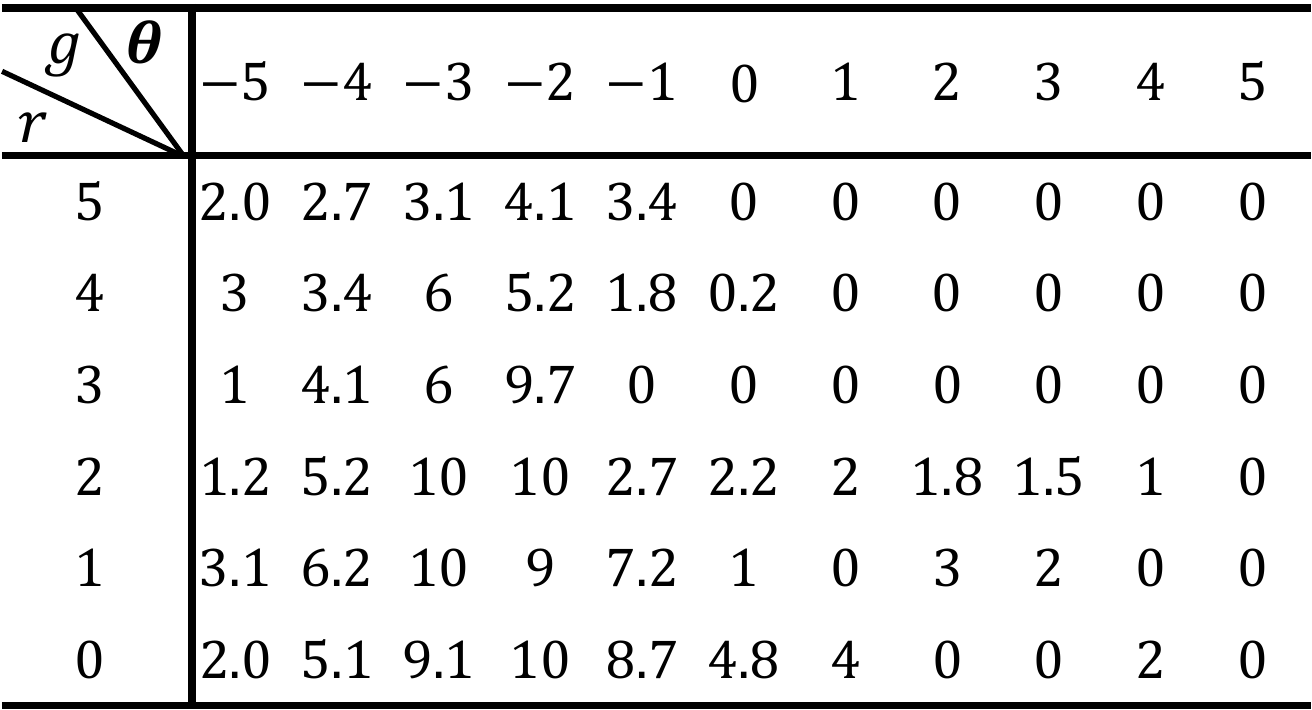}}
  \subfigure[Information gain curves]{\includegraphics[width=1.65in]{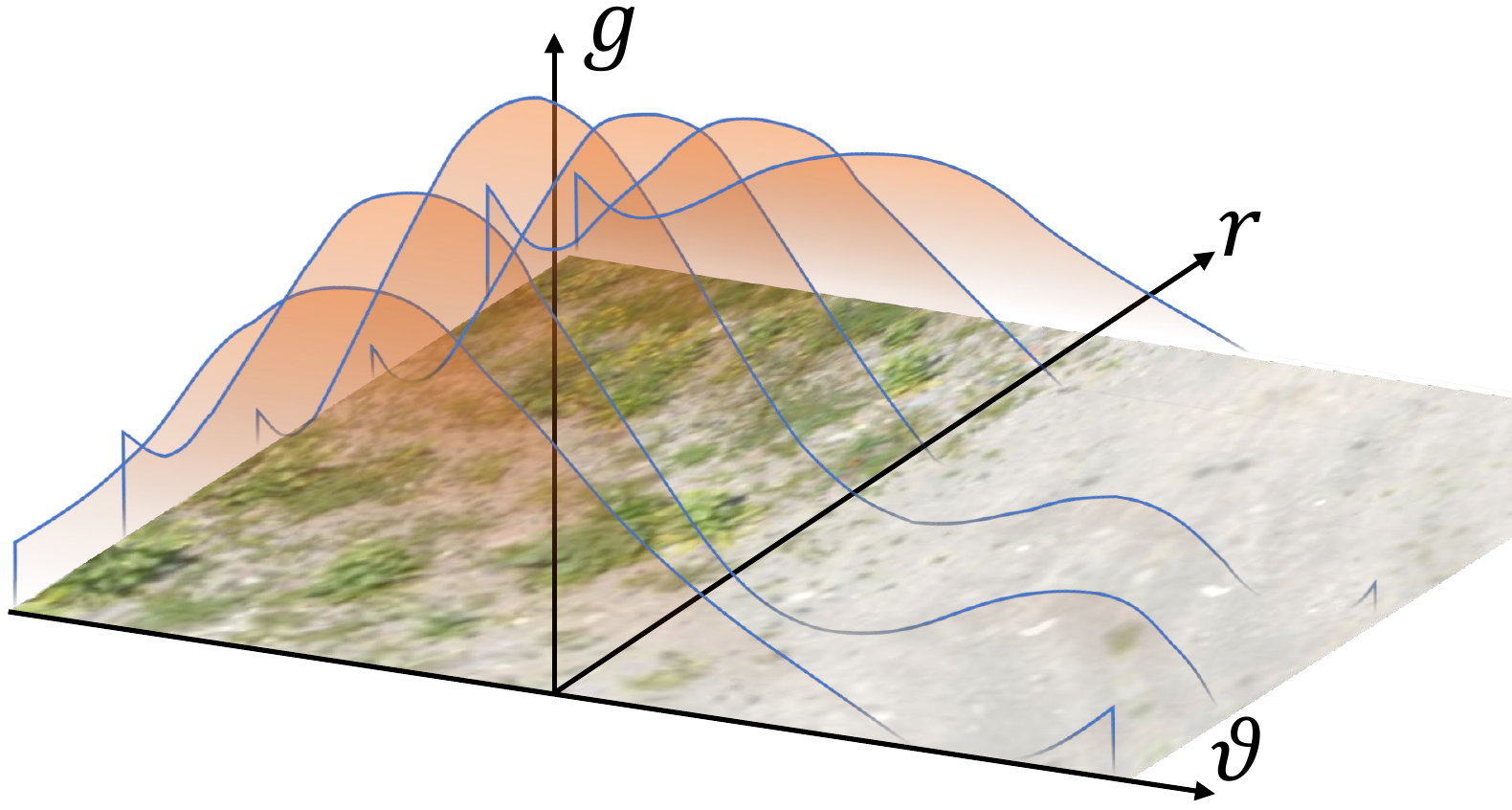}}
  \caption{Illustration of information gain curve fitting. (a) The yellow squares denote the sample points $\bm{p}_{i,j}$. The example environment includes feature-rich grass (left area) and featureless ground (right area). (b) The sample points are transferred into polar coordinates with $\theta$ and $r$ as horizontal and vertical axes. {The units of $r$ and $\theta$ axes are $m$ and $rad$, respectively.}  {(c)} The information gain of the sample points. (d) Each curve is fitted as a polynomial with degree $n$=6 from the sample points along corresponding sampling circle.}
  \label{fig:curve fitting}
  \vspace{-0.2cm}  
\end{figure}

The least squares method is utilized to solve $\bm{A}_i$, and we have
\begin{equation}\label{eq:optimal parameter}
  \setlength{\abovedisplayskip}{3pt}
  \setlength{\belowdisplayskip}{3pt}  
  \bm{A}_i = ({\bm{\varTheta}_i}^T{\bm{\varTheta}_i})^{-1}{\bm{\varTheta}_i}^T \bm{G}_i.
\end{equation}
The solution of the weight parameters $\bm{A}_i$ is analytical, contributing to computationally efficient. 
By fitting curves for each sample circle $C_i$, $N_{AP}$ polynomial functions are obtained for representing the information gain about the local environment.
Fig.\ref{fig:curve fitting} illustrates the process of information curves fitting.

The best point $\vartheta_i^b$ with the maximal information gain of the single polynomial function $f_i^g$ is obtained by
\begin{equation}\label{eq:maximal information gain for f_i^g}
  \setlength{\abovedisplayskip}{3pt}
  \setlength{\belowdisplayskip}{3pt}  
  \begin{aligned}
    \vartheta_i^b = \underset{\vartheta}{\arg \max } \  f_i^g(\vartheta).
  \end{aligned}
\end{equation}
The solution of \eqref{eq:maximal information gain for f_i^g} is obtained with the gradient descent method. 
Note that $f_i^g(\vartheta)$ is a non-convex function, suffering from the local minimal problem. 
Fortunately, according to the processes in Sections \ref{Generation of sample points} and \ref{Establishment of information gain function}, the best sample point $^{s}\bm{{p}}_{i}^{b}$ is easily obtained by comparing all sample points along $C_i$.
$^{s}\bm{{p}}_{i}^{b}$ provides a reliable initial value for numerical optimization of \eqref{eq:maximal information gain for f_i^g}, 
where the $\vartheta$ value of $^{s}\bm{{p}}_{i}^{b}$ is obtained from (9).
The best sample point $^{s}\bm{{p}}_{i}^{b}$ for $C_i$ and the best point ${{\bm{p}}}_{i}^{b}$ for $f_i^g$ are demonstrated in Fig.\ref{fig:planner}. 

\textcolor{red}{}
An information gain model $F$ is formulated to obtain the global best point $\bar{\bm{p}}^{b}$ with in all polynomial functions about the local environment.
The formulation is given as
\begin{equation}\label{eq:information gain formulation}
  \setlength{\abovedisplayskip}{3pt}
  \setlength{\belowdisplayskip}{3pt}  
  \begin{aligned}
    F(\bm{{\chi}}^\mathrm{wc}, \mathcal{M}, \bm{p}) = \sum_{i = 0}^{N_{AP}}{\delta(\bm{p}) \cdot f_i^g(\vartheta)}
  \end{aligned}
\end{equation}
where $\delta$ is a unit impulse function\textemdash the value of $\delta(\bm{p})$ is 1 if the point $\bm{p}$ is at the $i$th sample circle $C_i$, otherwise $\delta(\bm{p})$ is 0.

\textcolor{red}{}The problem of obtaining the global best point $\bar{\bm{p}}^{b}$ with maximal information gain is formulated as  
\begin{equation}\label{eq:maximal information gain for F}
  \setlength{\abovedisplayskip}{3pt}
  \setlength{\belowdisplayskip}{3pt}  
  \begin{aligned}
    \bar{\bm{p}}^{b} := \underset{\bm{p}}{\arg \max } \  F(\bm{{\chi}}^\mathrm{wc}, \mathcal{M}, \bm{p}).
  \end{aligned}
\end{equation}
The optimization problem is solved by selecting the point with maximal information gain in $\{{\bm{p}}_{0}^{b}, {\bm{p}}_{1}^{b}, ..., {\bm{p}}_{i}^{b}, ..., {\bm{p}}_{N_{AP}}^{b}\}$ as the global best point $\bar{\bm{p}}^{b}$. 
The global best point is shown as the solid blue circle in Fig.\ref{fig:planner}.

\vspace{-1em}
\subsection{{Receding Horizon Optimization}}

The optimal point $\bar{\bm{p}}^{b}$ is obtained by evaluating the local environment information gain at the current robot state. This is also known as single-step optimization.
However, the executed point sequence obtained by sequential single-step optimizations is discontinuous and disordered, resulting in unnecessary or uninformative motion. 
Because single-step optimization obtains the optimal point with only considering the current state and cannot involve future states to perform overall optimization for the future point sequence.
Therefore, optimizing the point sequence in a horizontal sliding window is needed for robust and continuous motion planning of the camera. 
{Then, a method based on the receding horizon optimization is developed to maximize the environmental information of future point sequence and minimize motion smoothness cost between neighbor points in the sequence.}

Because the trajectory of robot base is a prior given, the future robotic positions in $L$ steps, defined as $\{\bm{{\chi}}^\mathrm{wc}_{k+1}, ..., \bm{{\chi}}^\mathrm{wc}_{k+L}\}$, are available from the trajectory. Note that even if the trajectory is unknown, the robot can predict the future positions by a constant velocity motion model. 
We assume that the voxel map $\mathcal{M}$ maintains the same during the time of the horizontal sliding window. 
Then, the optimization problem including the information gain and motion smoothness is defined as 
\begin{equation}\label{eq:MPC objective function}
  \setlength{\abovedisplayskip}{3pt}
  \setlength{\belowdisplayskip}{3pt}  
  \begin{aligned}
    {\hat{\bm{p}}}^{b}_{k+1:k+L} = & \underset{\tilde{\bm{p}}_{k+1:k+L}}{\arg \min } \  -\lambda_{info} \mathcal{J}_{info} + \lambda_{smo} \mathcal{J}_{smo} \\
  \end{aligned}
\end{equation}
where $\tilde{\bm{p}}_{k+1:k+L}$ denotes the view-landing-points to be optimized in the horizontal sliding window; $\mathcal{J}_{info}$ denotes the information gain; $\mathcal{J}_{smo}$ denotes the smoothness cost penalizing trajectory discontinuity; 
{\mbox{$\lambda_{info}$} and \mbox{$\lambda_{smo}$} are weight coefficients to balance the two terms \mbox{$\mathcal{J}_{info}$} and \mbox{$\mathcal{J}_{smo}$} - the more significant value of one coefficient than the other, the more concerned about the related term. The values of \mbox{$\lambda_{info}$} and \mbox{$\mathcal{J}_{smo}$} were set empirically.}
From \eqref{eq:information gain formulation}, the information gain term $\mathcal{J}_{info}$ is defined as 
\begin{equation}\label{information cost term}
  \setlength{\abovedisplayskip}{3pt}
  \setlength{\belowdisplayskip}{3pt}  
  \mathcal{J}_{info} = \sum_{t = k+1}^{k+L}{(\frac{1}{2}{F(\bm{{\chi}}^\mathrm{wc}_{t}, \mathcal{M}, \tilde{\bm{p}}_{t})}^2)}
\end{equation}
where $\tilde{\bm{p}}_{t}$ denotes the point to be optimized at time $t$.
{Further, to avoid unnecessary or discontinuous motion, the smoothness cost term $\mathcal{J}_{smo}$ is given by evaluating the variation between neighbor points. The displacement vector between $\tilde{\bm{p}}_{t} $ and $ \tilde{\bm{p}}_{t-1}$ is calculated as $\Delta \tilde{\bm{p}}_{t} = \tilde{\bm{p}}_{t} - \tilde{\bm{p}}_{t-1}$. And then, the smoothness cost term is defined as}
\begin{equation}\label{smoothness cost term}
  \setlength{\abovedisplayskip}{3pt}
  \setlength{\belowdisplayskip}{3pt}  
  \mathcal{J}_{smo} = \sum_{t=k+1}^{k+L}(\Delta \tilde{\bm{p}}_{t+1} - \Delta \tilde{\bm{p}}_{t})^2,
\end{equation}
{this term constrains the neighbor displacement vectors in both the direction and length. The cost indicates the smoothness and distance distribution of $ \tilde{\bm{p}}_{t+1}, \tilde{\bm{p}}_{t}$, and $ \tilde{\bm{p}}_{t-1}$.}

{Since the complex information gain function in \mbox{\eqref{eq:cost candidate extend}} is converted to the formulation in \mbox{\eqref{eq:information gain formulation}} which is differentiable, the optimization problem \mbox{\eqref{eq:MPC objective function}} is solved by using the gradient descent method\mbox{\cite{boyd2004convex}}.
And the single-step optimized point $\bar{\bm{p}}^{b}_{t}$ by \mbox{\eqref{eq:maximal information gain for F}} is used as initial state. 
The $\tilde{\bm{p}}_{t}$ iterates as follows}
\begin{equation}\label{gradient descent}
  \setlength{\abovedisplayskip}{3pt}
  \setlength{\belowdisplayskip}{3pt}  
  \tilde{\bm{p}}_{t} := \tilde{\bm{p}}_{t} - (\lambda_{info} \cdot \frac{\partial \mathcal{J}_{info}}{\partial \tilde{\bm{p}}_{t}} + \lambda_{smo} \cdot \frac{\partial \mathcal{J}_{smo}}{\partial \tilde{\bm{p}}_{t}}).
\end{equation}
{The gradient of $\mathcal{J}_{info}$ with respect to $\tilde{\bm{p}}_{t}$ is calculated as}
\begin{equation}\label{gradient of information cost term}
  \setlength{\abovedisplayskip}{3pt}
  \setlength{\belowdisplayskip}{3pt}  
  \frac{\partial \mathcal{J}_{info}}{\partial \tilde{\bm{p}}_{t}} \Leftrightarrow \frac{\partial \mathcal{J}_{info}}{\partial \theta_{t}} \Leftrightarrow  \frac{\partial f_i^g}{\partial \theta_{t}} = \sum_{s=1}^{n} {a}_{i,s} \cdot  ({{\theta}_{i,t}})^{s-1}
\end{equation}
{where $\Leftrightarrow$ denotes logical equivalence; the first $\Leftrightarrow$ indicates that $\tilde{\bm{p}}_{t}$ and $\theta_{t}$ represent the same point; the second $\Leftrightarrow$ indicates that only one polynomial function $f_i^g$ in $F$ relates to the variable $\tilde{\bm{p}}_{t}$ when $\delta(\tilde{\bm{p}}_{t})=1$ according to \mbox{\eqref{eq:information gain formulation}}, because $f_i^g$ is fitting from the sampling circle $C_i$ where $\tilde{\bm{p}}_{t}$ lays in, and $C_i$ is determined by the distance between $\tilde{\bm{p}}_{t}$ and $\bm{{\chi}}^\mathrm{wc}_{t}$ through \mbox{\eqref{eq:generate sample}}. The gradient of $\mathcal{J}_{smo}$ with respect to $ \tilde{\bm{p}}_{t}$ is calculated as}
\begin{equation}\label{gradient of smoothness cost term}
  \setlength{\abovedisplayskip}{3pt}
  \setlength{\belowdisplayskip}{3pt}  
  \begin{aligned}
    \frac{\partial \mathcal{J}_{smo}}{\partial \tilde{\bm{p}}_{t}} 
    & =-4 *\left(\tilde{\bm{p}}_{t+1}-2 \cdot \tilde{\bm{p}}_{t}+\tilde{\bm{p}}_{t-1}\right).
  \end{aligned}
\end{equation}

{For receding horizon optimization, the optimizing variable is the point sequence $\tilde{\bm{p}}_{k+1:k+L}$, where each point iterates according to \mbox{\eqref{gradient descent}}. And the iteration step index of \mbox{\eqref{gradient descent}} is omitted for simplicity of statement.}
After iterating $\tilde{\bm{p}}_{k+1:k+L}$ until convergence, the optimal solution of the best point sequence is obtained and remarked as ${\hat{\bm{p}}}^{b}_{k+1:k+L}$. The solution balanced information gain and motion smoothness in the horizontal time window.
Finally, the first point ${\hat{\bm{p}}}^{b}_{k+1}$ in ${\hat{\bm{p}}}^{b}_{k+1:k+L}$ is selected as the next desired best point. 
The bottom tracking controller is then utilized to output control vector $u^*$ according to the inverse kinematics and drive the gimbal camera towards ${\hat{\bm{p}}}^{b}_{k+1}$.

{Further, the view direction may swing with the robot base in practical field environments when moving on rough terrains. The gimbal's bottom controller is used not only to track the best point but also to improve the localization failure problem caused by the view direction swing by controlling the gimbal's pitch. Moreover, the controller outputs control commands at a high frequency with 100Hz.} 

\vspace{-0.2cm}  
\section{Simulations and Experiments}\label{section simulation}

\subsection{Physical-engine simulation and experimental platforms}
\begin{figure}[htbp]
  \setlength{\abovecaptionskip}{0.cm}
  \vspace{-0.2cm}  
  \centering
  \subfigure[Gazebo simulated robot]{\includegraphics[width=1.7in]{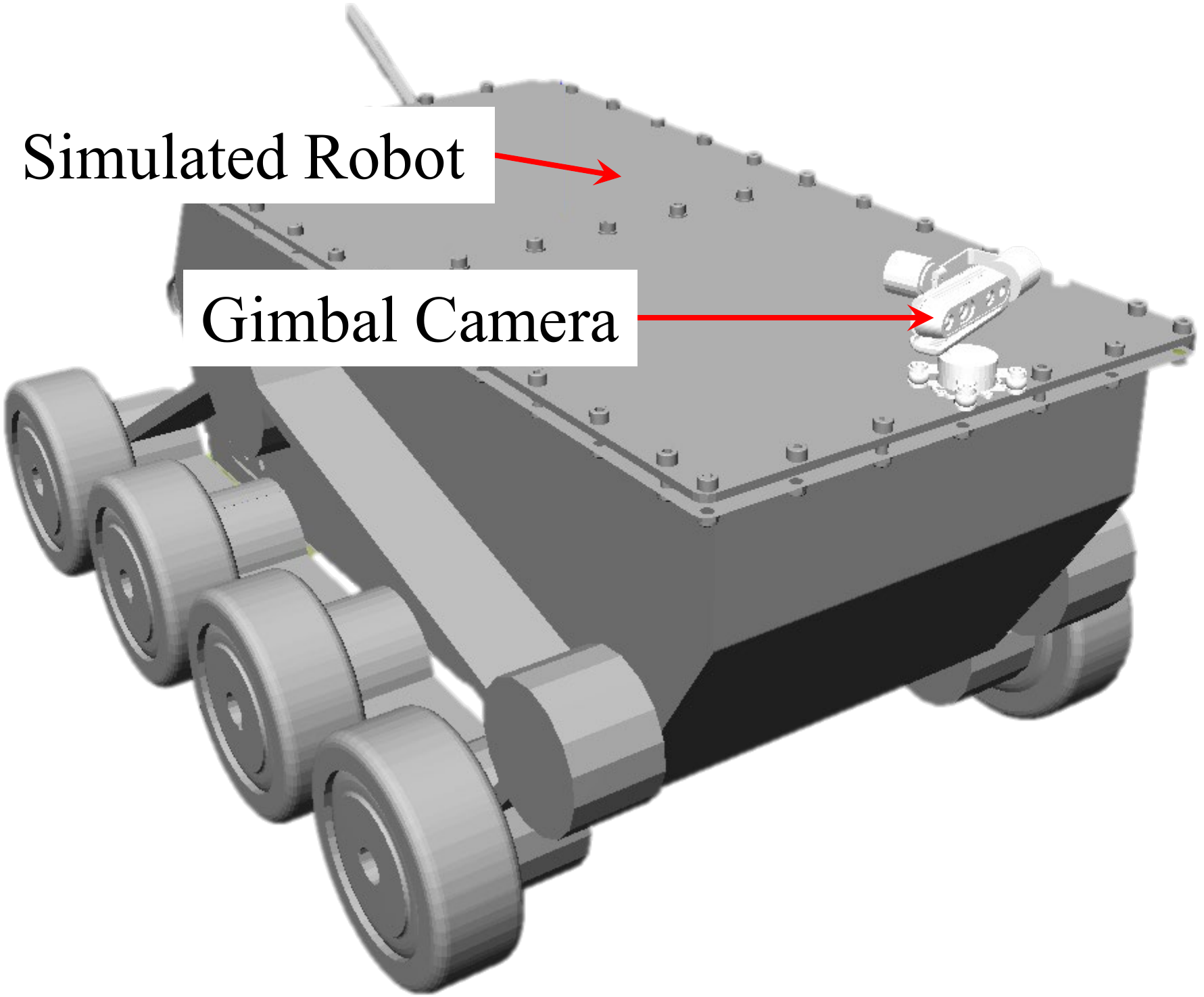}}
  \subfigure[Experimental terrain vehicle]{\includegraphics[width=1.7in]{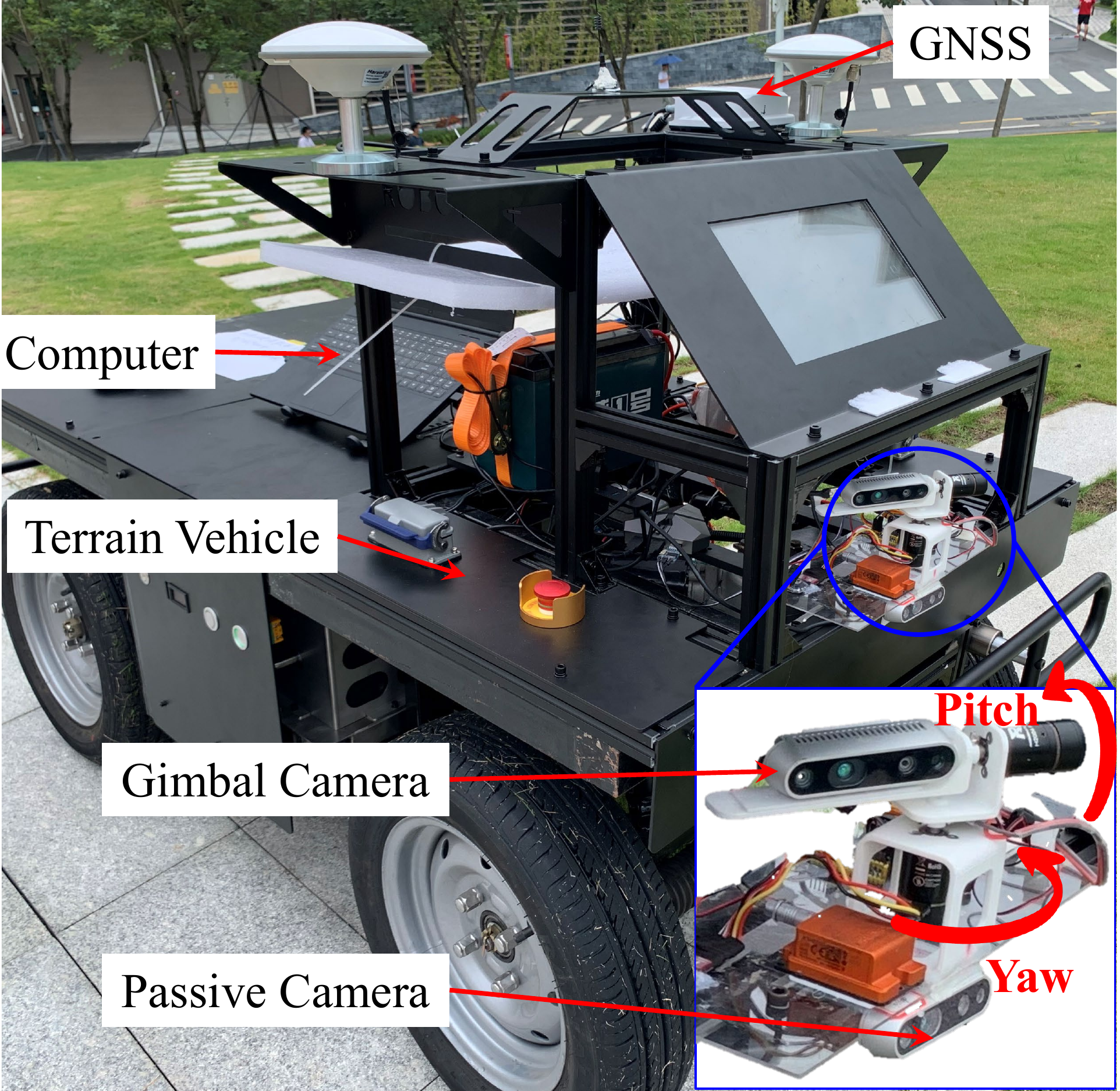}}
  \caption{{Simulation and experiment platforms.}}
  \label{fig:Robot Flatform}
  \vspace{-0.2cm}  
\end{figure}

Several simulations and experiments were performed to verify the proposed approach. 
We built a mobile platform in the physics engine-based simulator Gazebo, and utilized an experimental terrain vehicle, as shown in Fig.\ref{fig:Robot Flatform}. 
An RGB-D camera on the gimbal was equipped for perception in the simulations and experiments. Another RGB-D camera was fixed on the robot to compare the passive method. 
The gimbal is of two-axis, with the pitch and yaw angles being controllable. 
{A high-precision GNSS was used to provide ground truth in the experiments.
The parameters are listed in TABLE \mbox{\ref{table:parameters}}. $P_{min}$ was set 2$m$, which can exclude the evaluation of the features too close to the camera. }

\begin{table}[htbp]
  \vspace{-1em}
  \caption{Experiment parameters }
  \centering
  \begin{tabular}{cc|cc|cc}
  \Xhline{1pt}
  Camera FOV        & $69^\circ \times 42^\circ$ & $\Delta_d$            & 0.4 m & n                    & 6    \\
  Camera resolution & $640 \times 480$ &  ${\Delta}_{\theta}$ & $20^\circ$   & L                             & 6    \\
  Max. sensor range & 10 m      & {$P_{min}$}                    & 2 m  & {$\lambda {info}$}   & 1    \\
  Max. velocity       &  20 km/h      & $N_{AP}$             & 10  & {$\lambda_{smo}$} & 0.12 \\   
  Voxel size             &  0.4 m      & $N_{SP}$                     & 18  &  \\   \Xhline{1pt}
  \end{tabular}
  \label{table:parameters}
\end{table}

\subsection{Evaluation of information mapping method}

\begin{figure}[htbp]
  \setlength{\abovecaptionskip}{0.cm}
  \vspace{-0.2cm}  
  \centering
  \subfigure[Simulation environment]{\includegraphics[width=1.9in]{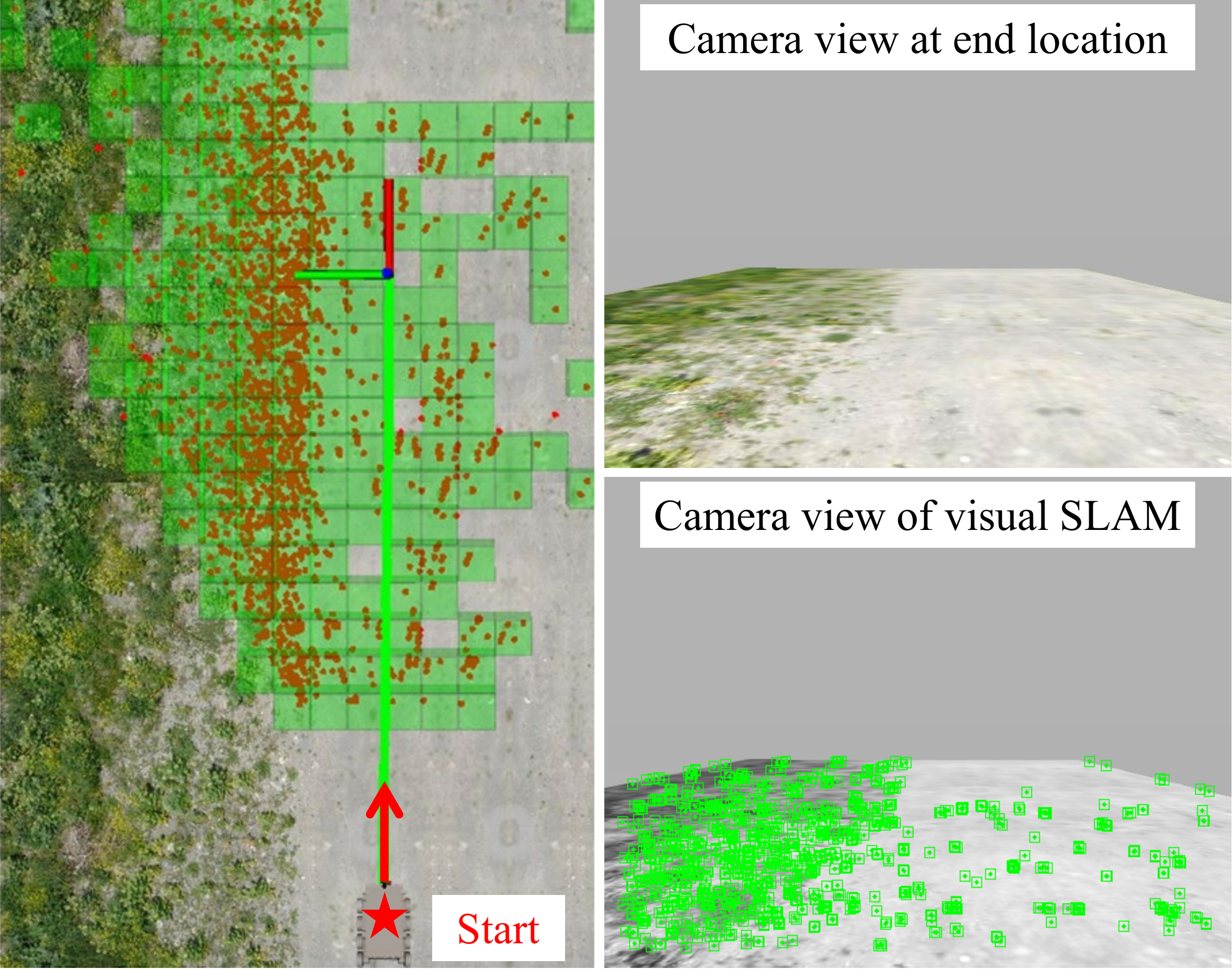}}
  \subfigure[Information map layers]{\includegraphics[width=1.5in]{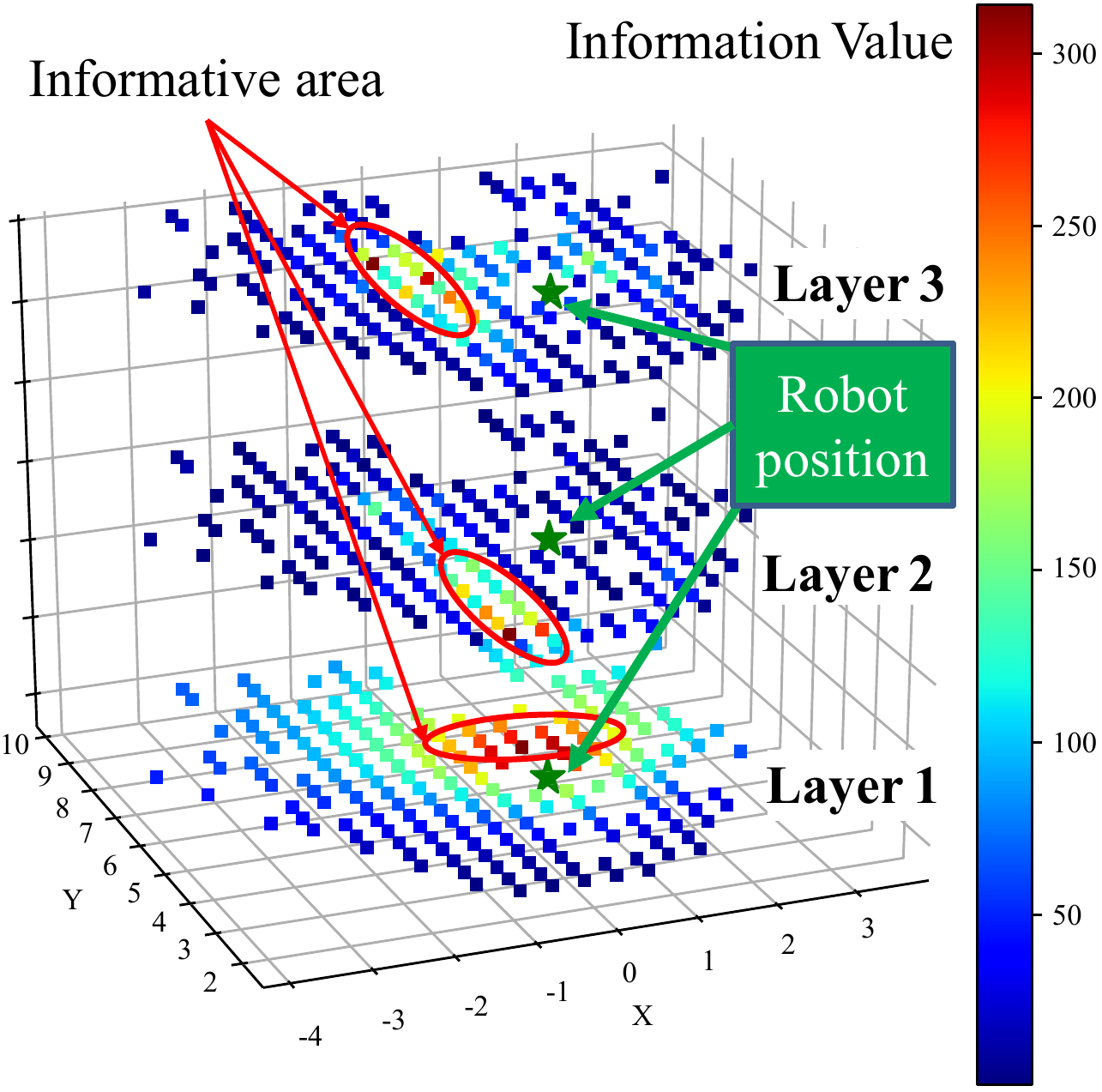}}
  \caption{Illustration of information mapping. (a) The red dots denote the built map points, the green line denotes the robotic trajectory, and the green voxels denote the voxel map made from the map points. (b) Layer 1, Layer 2, and Layer 3 denote the maps for calculating Fisher information, feature distribution information and weighted Fisher information, respectively. The heat map represents the information value at each voxel.}
  \label{fig:information map}
  \vspace{-0.4cm}  
\end{figure}

We first performed a simulation to evaluate the proposed information mapping method. 
Figure \ref{fig:information map}(a) illustrates the top view of an example environment. The robot's left side is texture-rich grass, and the right side is textureless ground. The robot with the fixed camera traveled along the trajectory shown as the green line in Fig.\ref{fig:information map}(a). 
As shown in Fig.\ref{fig:information map}(b), 
the information maps are built by different information formulations, i.e., the Fisher information in \eqref{eq:trace of fisher information without rotation}, the feature distribution information in \eqref{eq:feature distribution}, and the distribution-weighted Fisher information in \eqref{eq:total_information}, but without considering the exploration factor. 
In the Fisher information map (Layer 1), the textureless ground near the robot also incorrectly provides high information values. 
Because the voxel focuses on the occupancy, the textureless ground can provide sparse feature to occupy the voxel. This results in the left texture-rich grass having the same Fisher information with the right textureless ground.
The feature distribution information map (Layer 2) makes the visited area at the bottom left of Layer 2 to be informative area.
The distribution-weighted Fisher information map (Layer 3) complements the Fisher information with the feature distribution information. 
It is seen that the information representation provided in Layer 3 is more accurate compared to the other two methods, because
the area takes both the Fisher information and the feature distribution information into consideration according to \eqref{eq:total_information}. 

{Since the Fisher information in the above simulations considered the uncertainty of the voxel position by the covariance $\bf{Q}$, we additionally performed simulations to verify the effects of feature uncertainty on the information mapping method. The feature uncertainty is represented by $\bf{Q^*}=diag({\delta_p}^2,{\delta_p}^2,{\delta_z}^2)$ where ${\delta_p}^2=0.25$ denotes the feature points' pixel error and $\delta_z = 1.425 \times 10^{-6}z^2(mm^2)$ depends on depth measurement\mbox{\cite{proenca2018probabilistic}}. 
The results show that the information distribution with and without adding feature uncertainty $\bf{Q^*}$ are consistent. 
In addition, the detection stability of 2D feature points can be one source of feature uncertainty. 
According to our tests, the whole feature point map output from SLAM can distinguish feature distribution differences well and guide for view planning, even though the detected 2D features are not exactly the same in each operation even in the same environment.
Therefore, the feature uncertainty has minor effects on the information distribution and does not affect the active view planning for perception in this work. 
}

\begin{figure}[htbp]
  \setlength{\abovecaptionskip}{0.cm}
  \vspace{-0.2cm}  
  \centering
  \subfigure[View-landing-points planned by different maps]{\includegraphics[height=1.8in]{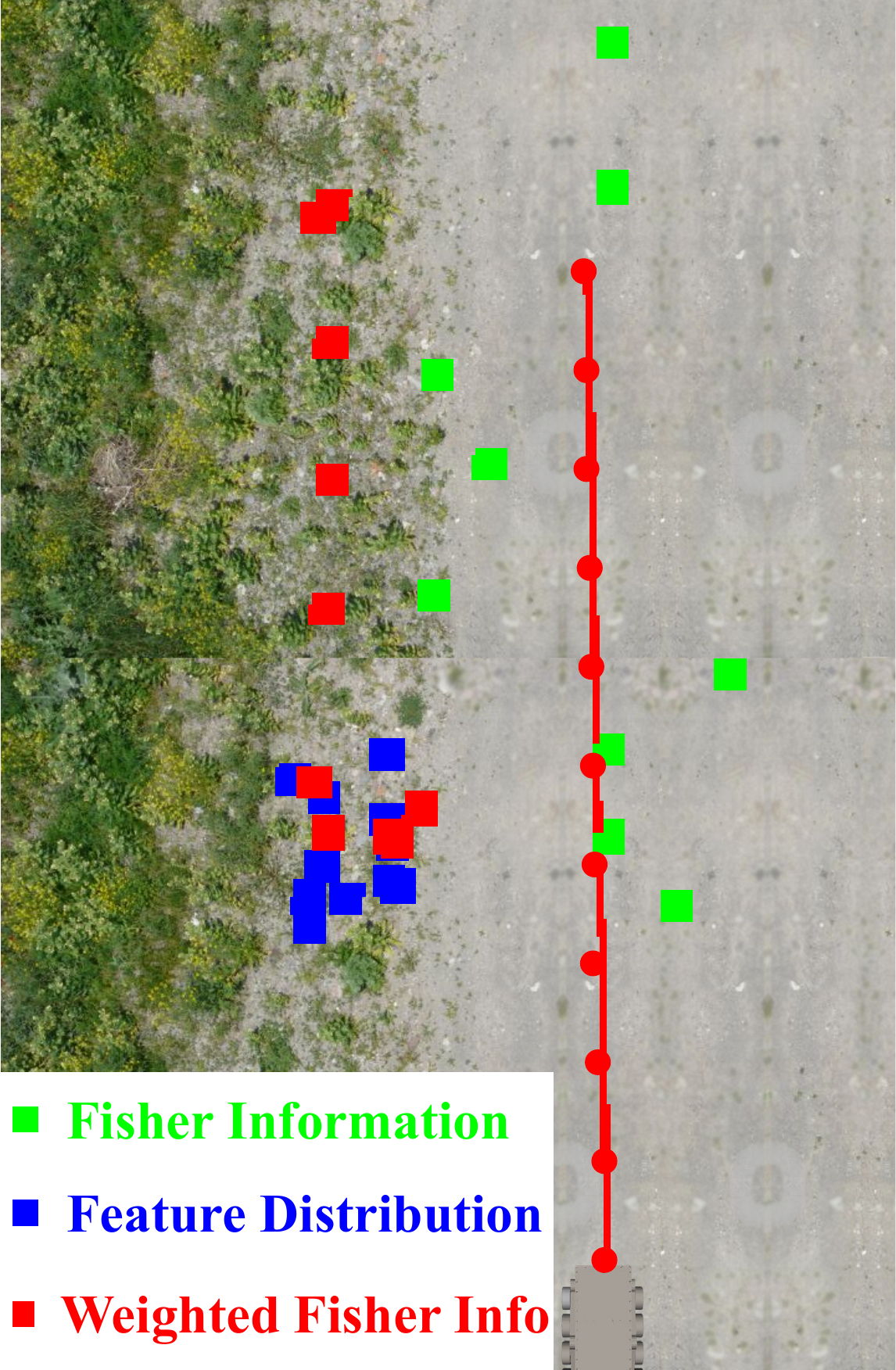}}
  \subfigure[Estimation errors]{\includegraphics[height=1.8in]{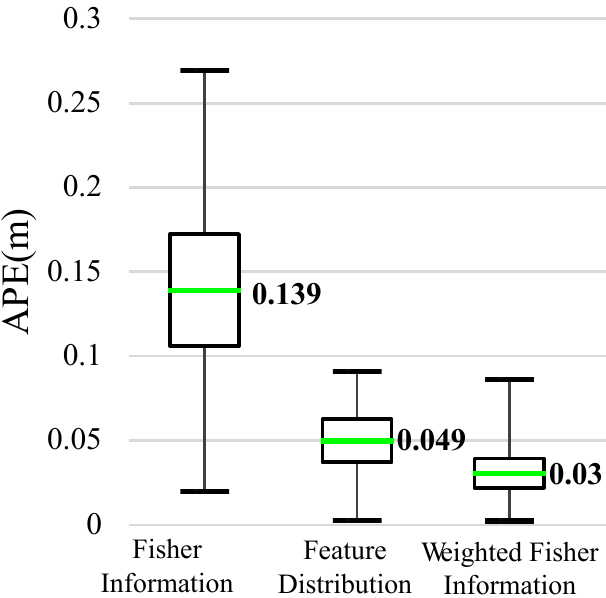}}
  \caption{ Evaluation Simulation results of localization accuracy under different information mapping methods. (a) The red line denotes the robotic trajectory, the red dots denote the trajectory points at different timestamps. The colored squares represent the best view-landing-points under different information maps. (b) The mean APE is denoted by a green line, while the STD of the estimation error is denoted by a black box.}
  \label{fig:Mapping method evaluation}
\end{figure}

To evaluate the performance of different information mapping methods on the localization accuracy, we designed a simulation to make the robot travel 6 meters along a straight line. 
Without loss of generality, we utilized the regular sampling method in Section \ref{section4}.A to plan the camera view (i.e., view-landing-point). 
The planned view-landing-points were illustrated in Fig.\ref{fig:Mapping method evaluation}(a) under different information maps. 
The absolute pose error (APE) between the estimated trajectory and the ground truth trajectory \cite{grupp2017evo} is used to quantify localization accuracy {after repeating the simulation five times}, as shown in Fig.\ref{fig:Mapping method evaluation}(b). 
It is seen that the Fisher information map exhibited maximum estimation error because the planned view-landing-points(the green squares in Fig.\ref{fig:Mapping method evaluation}(a)) scattered on the textureless ground. 
Indicated by blue squares in Fig.\ref{fig:Mapping method evaluation}(a), the view-landing-points planned under the feature distribution information map concentrated on the known area that has been visited, thus exhibiting smaller tracking errors than the Fisher information map. 
However, using only the feature distribution information map leads to exploration degeneration, and makes the camera towards to the back of the robot and to be unfavorable for feature tracking in this work.
Indicated by the red squares in Fig.\ref{fig:Mapping method evaluation}(a), the view-landing-points of the distribution-weighted Fisher information map laid on the texture-rich grass. Hence, the estimation error performed well with the smallest error. Moreover, compared with the feature distribution information map, the distribution-weighted Fisher information map made the view-landing-points be distributed at the upper right of the robot pose. 
Because the points closer to the robot have higher Fisher information according to the definition in \eqref{eq:fisher information jacobian}, which makes the fusion of Fisher information and the distribution information alleviate the degree of exploration degeneration. 
Therefore, the distribution-weighted Fisher information map is more suitable for view planning thanks to considering both the quality and distribution of the environmental features. 

\begin{figure}[htbp]
  \setlength{\abovecaptionskip}{0.cm} 
  \centering
  \subfigure[Polynomial regression models]{\includegraphics[width=1.7in]{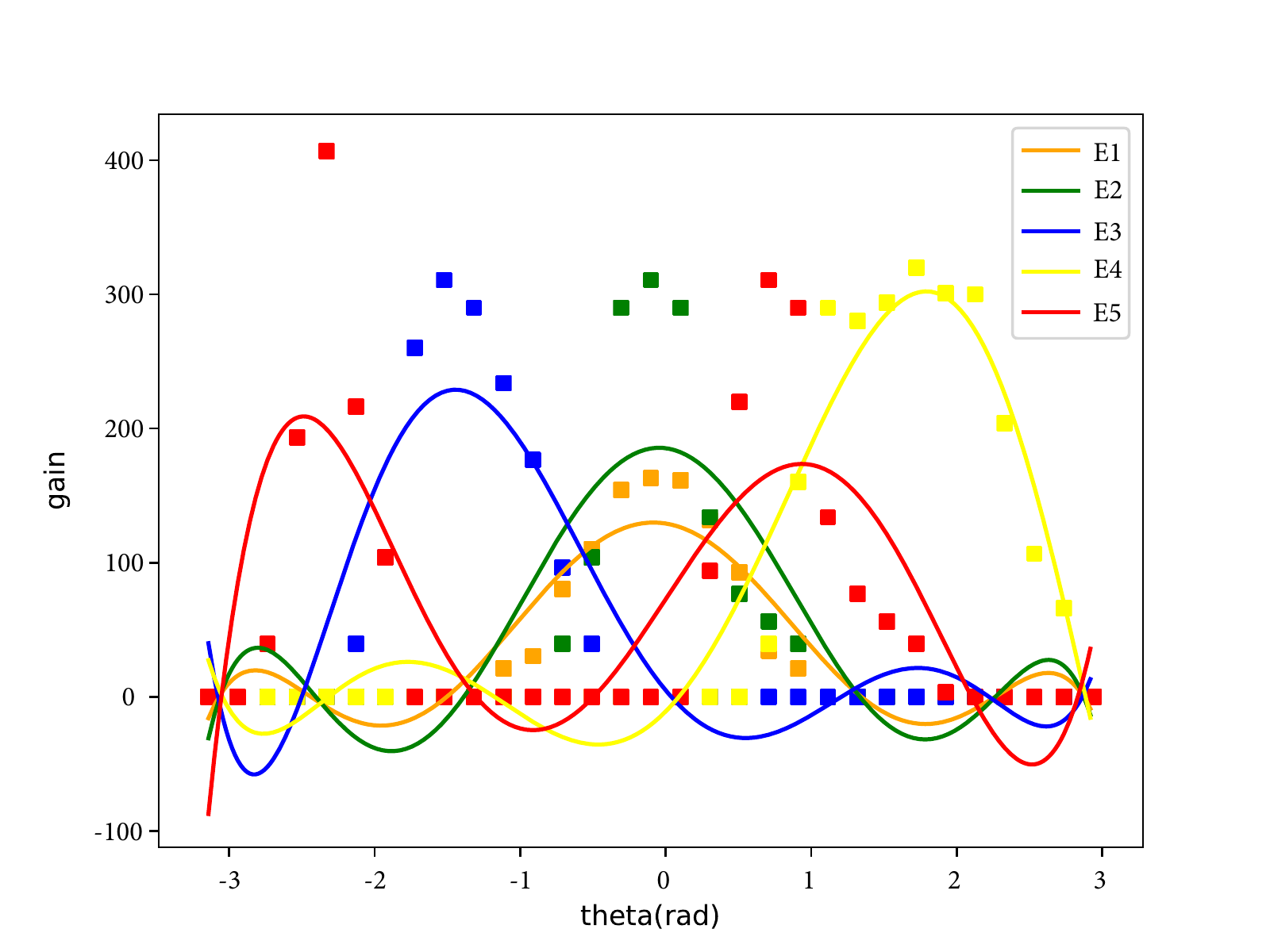}}
  \subfigure[Regression errors]{\raisebox{0.25\height}{\includegraphics[width=1.7in]{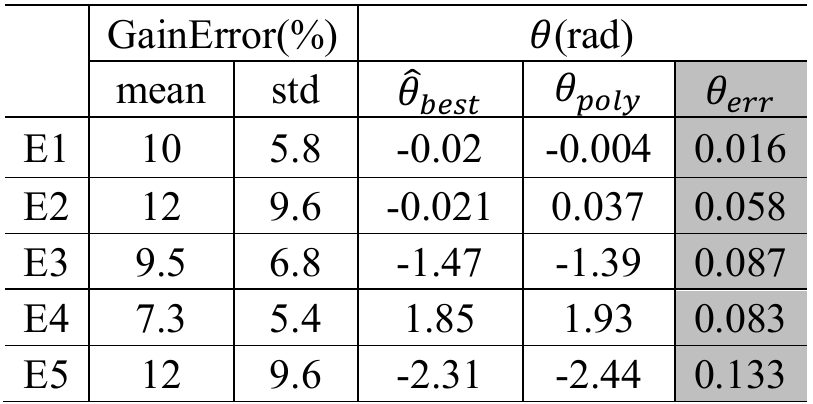}}}
  \caption{ {Evaluation of the polynomial regression accuracy. }}
  \label{fig:Polynomial fitting evaluation}
  \vspace{-0.3cm}  
\end{figure}
{We have collected data at representative locations (i.e., E1, E2, E3, E4, and E5) to evaluate the accuracy of polynomial regression. 
Each of the obtained regression models indicated by coloured curves in Fig.\mbox{\ref{fig:Polynomial fitting evaluation}}(a) is based on the points sampled along a circle around the robot location with the interval angle $\Delta_{\theta}=0.2 rad$. The information gain at each sample point is calculated by \mbox{\eqref{eq:cost candidate extend}} and indicated by a coloured small square.
In Fig.\mbox{\ref{fig:Polynomial fitting evaluation}}(b), the second and third columns denote the gain error between the ground truth gain and the polynomial regression gain of the sampled points, respectively.
Moreover, our work concentrates on the accuracy of the angle $\theta$ rather than the information gain because the planner finally uses the angle $\theta$ with the maximum information gain as the best view direction. 
The angle error $\theta_{err}$ between ${\hat{\theta}}_{best}$ and $\theta_{poly}$ is used to measure the regression accuracy. ${\hat{\theta}}_{best}$ denotes the true angle with maximum information gain, and the angle is obtained by dense sampling with the interval $\Delta_{\theta}=0.01 rad$. $\theta_{poly}$ denotes the maximum information angle obtained from the regression polynomial by the global optimization.
From Fig.\mbox{\ref{fig:Polynomial fitting evaluation}}(b), the $\theta_{err}$ of the polynomial curves regressed on data of E1, E2, E3, E4, and E5 are with a small value below 0.133 $rad$.
From Fig.\mbox{\ref{fig:Polynomial fitting evaluation}}(a), it is noted that the optimal angle obtained from the polynomial regression still stays near the area with high information, even though there exists regression error. This makes the localization accuracy improved by looking to the informative area.
Furthermore, according to our tests on the regression speed and accuracy performance, the time cost decreases and the error increase when the sample interval grew. }

\vspace{-1em}
\subsection{Evaluation of camera view planning algorithm}

The proposed camera view planning approach for robust visual SLAM includes information mapping and camera view planner IGLOV.
The information mapping module has been verified in the previous section; therefore, we further designed several simulations to evaluate the performance of IGLOV planner.
We compared the proposed IGLOV planner with several existing planning methods, including the passive method (PAS), uniform sampling in view space (USV)\cite{zeng2020pc}, Monte Carlo sampling in task space (MST)\cite{palomeras2018autonomous}, regular sampling considering degeneration in task space (RSDT). 
For fair comparison, the previous weighted-Fisher information mapping method was used for these camera view planning approaches.
PAS method fixed the camera on the robot base and did not change the viewing direction.
USV method sampled ten views uniformly in $[-\pi, \pi]$ of yaw at each timestamp and evaluated each view to find the maximal information view.
MST method, also called random sampling method in \cite{palomeras2018autonomous}, sampled 500 points around the robot within 5m at each timestamp and evaluated each point to find the one with maximal information. USV and MST evaluate the information according to \eqref{eq:total_information}.
RSDT method sampled points according to \eqref{eq:generate sample} and selected the sample point with maximal information gain as the best view-landing-point according to \eqref{eq:cost candidate extend}. RSDT considered the consistency between the view and the motion direction but without applying the polynomial regression and receding horizon optimization.

\begin{figure}[htbp]
  \setlength{\abovecaptionskip}{0.cm} 
  \vspace{-0.2cm}  
  \centering
  \subfigure[Simulation environment]{\includegraphics[height=1.8in]{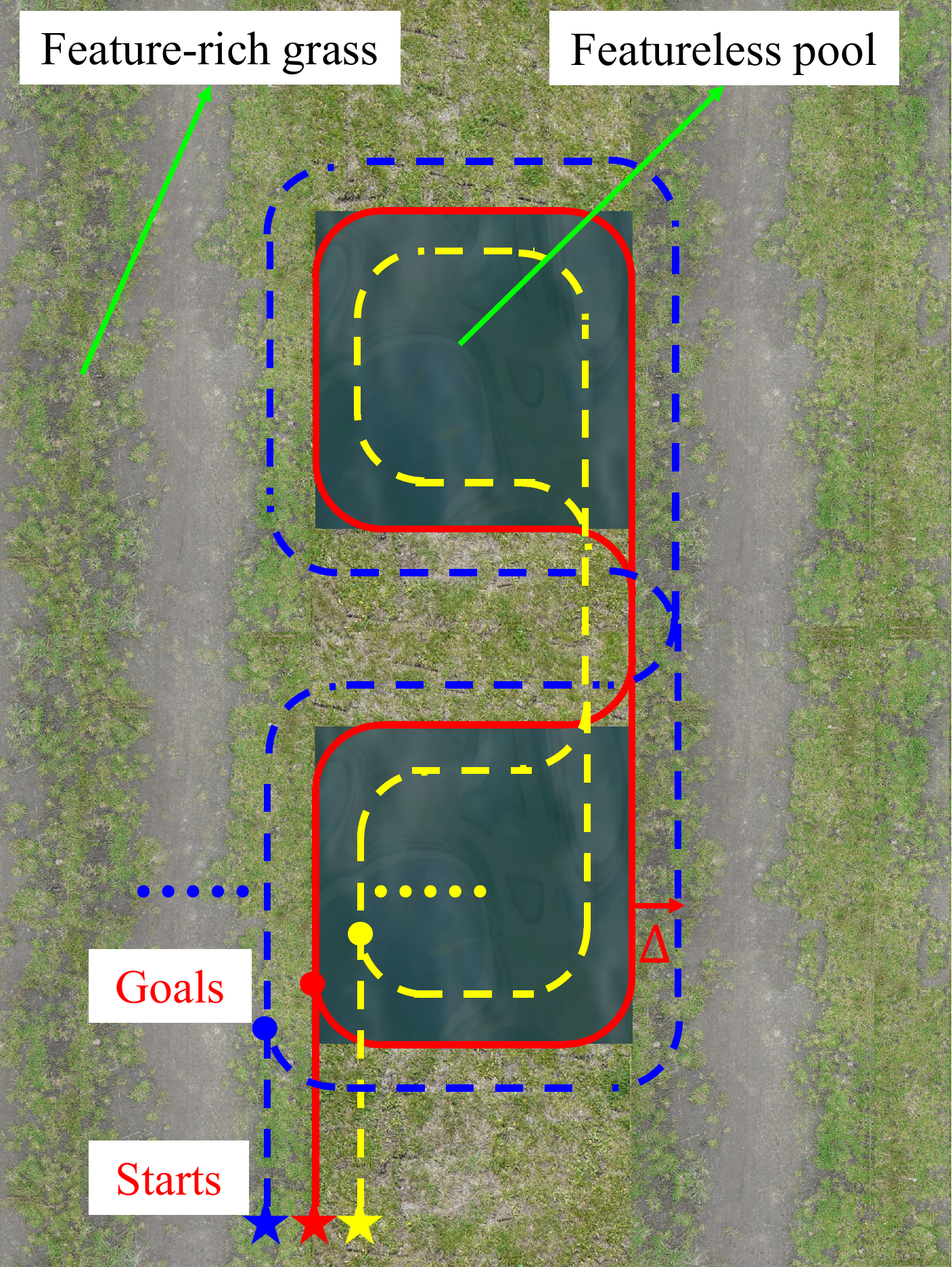}}
  \subfigure[Estimated trajectories]{\includegraphics[height=1.8in]{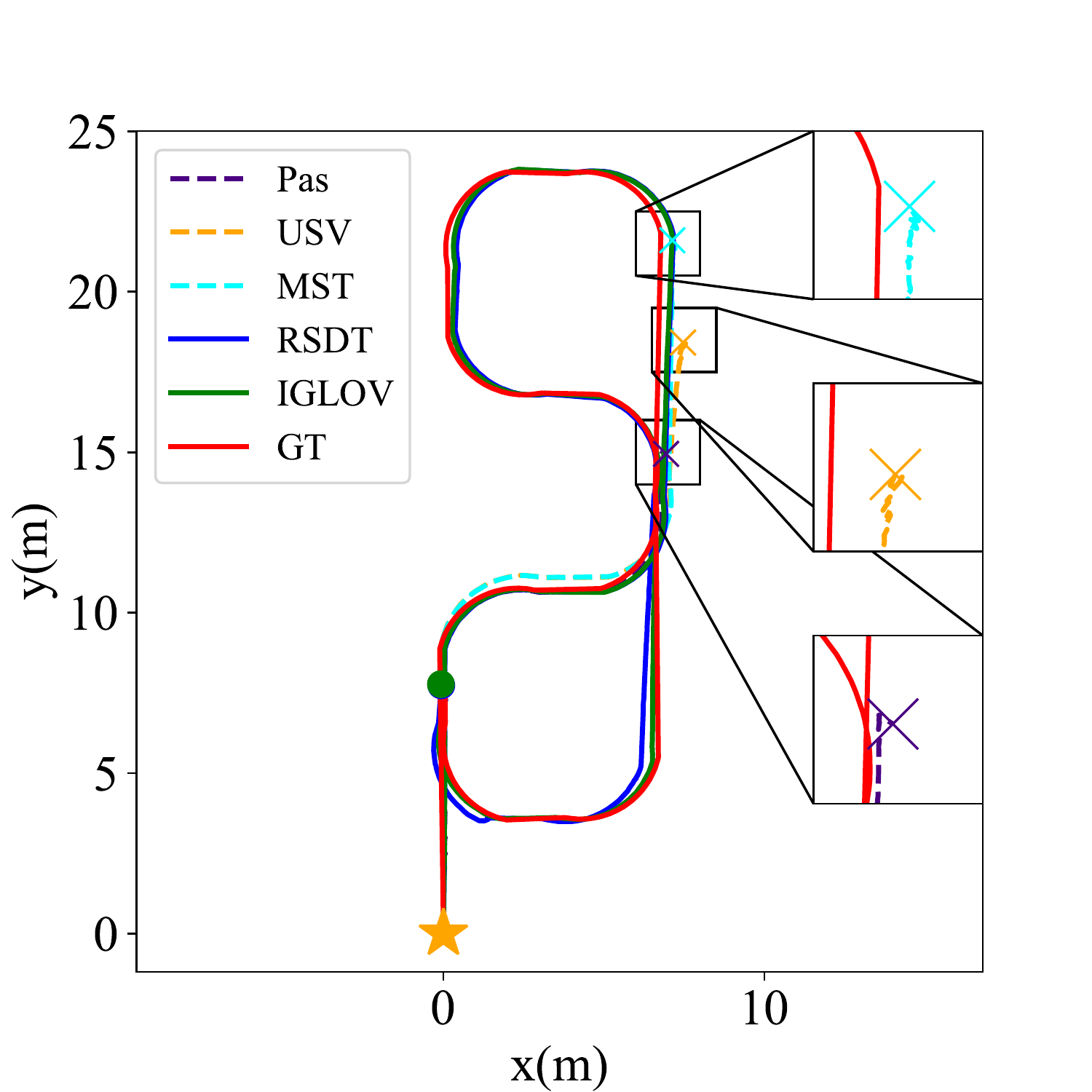}}
  \caption{ Simulation 1 and the results under different methods. (a) Two featureless pools lie in a grass plane, and the trajectories with different colors represent various offsets from the baseline trajectory. (b) Estimated trajectories under different planning methods at offset $\Delta = -0.6$. GT is the ground truth indicated by the solid red trajectory. The orange star marker is the start point of the trajectories. The circular markers `$\bullet$' with different colors represent the endpoint of a successful tracking, while the `$\times$' markers represent the endpoint of a failed tracking. 
  }
  \label{fig:simulation environment}
  \vspace{-0.2cm}  
\end{figure}

The first scene simulated a wild environment, as shown in Fig.\ref{fig:simulation environment}(a). The red curve denotes a preplanned baseline trajectory of the robot; the blue and yellow dot curves are obtained by transforming the baseline trajectory with a positive offset $\Delta > 0$ and negative offset $\Delta < 0$, respectively. Positive offset zooms the trajectory to the feature-rich area, while negative offset shrinks the trajectory to the featureless area. The simulated robot moved along five trajectories with $\Delta = \{0.6, 0.3, 0, -0.3, -0.6\}$ meters. The simulation also investigated how the proportion of featureless areas affects localization accuracy. 

\begin{figure}[htbp]
  \setlength{\abovecaptionskip}{0.cm} 
  \vspace{-0.2cm}  
  \centerline{\includegraphics[width=3.5in]{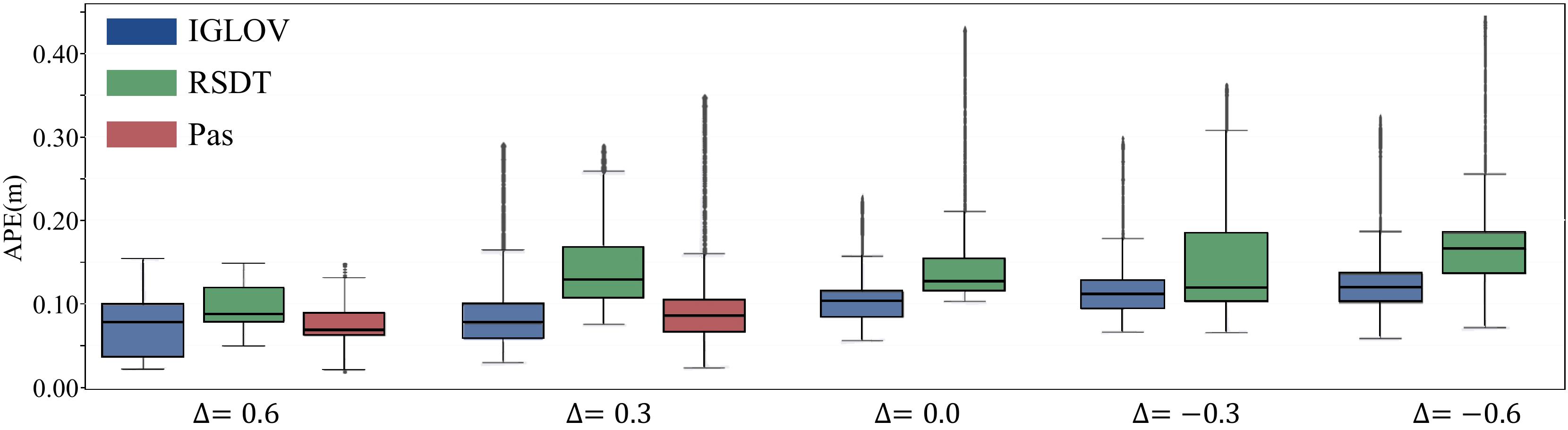}}
  \caption{Estimation errors under different planning methods with various offsets $\Delta$. The results of USV and MST are not presented due to the estimation failures.}
  \label{fig:ATE of simulation exeriments}
  \vspace{-0.2cm}  
\end{figure}

\begin{figure}[htbp]
  \setlength{\abovecaptionskip}{0.cm} 
  \vspace{-0.2cm}  
  \setlength{\abovecaptionskip}{0.cm}
  \centerline{\includegraphics[width=3.2in]{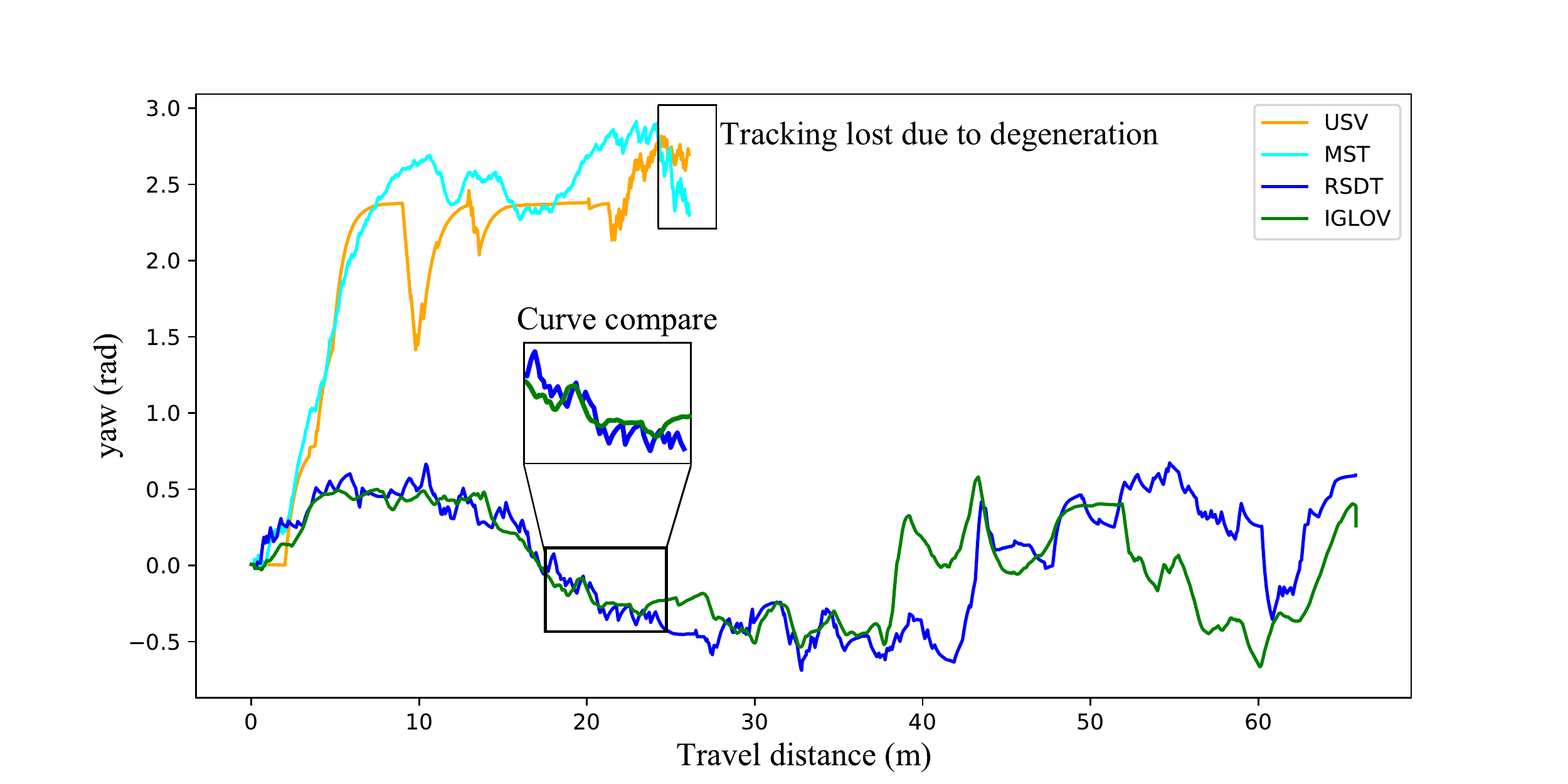}}
  \caption{Gimbal's yaw angle curve versus travel distance at the simulation with $\Delta = -0.6$. }
	\label{fig:yaw_angle}
  \vspace{-0.6cm}  
\end{figure}

As shown in Fig.\ref{fig:ATE of simulation exeriments}, the PAS method localized well with positive $\Delta$ thanks to the high proportion of featurerich areas in FOV. 
The USV and MST failed to localize the robot in all simulations. 
The gimbal's yaw angle curves for the simulation with $\Delta = -0.6$ in Fig.\ref{fig:yaw_angle} demonstrated the reason of localization failure. 
The yaw angles in orange and cyan curve reach a high value about 3.0rad after traveled 23m, which means the gimbal camera looked towards the robot's back by using the USV and MST methods; this resulted in the exploration degeneration and loss of feature tracking. The details can be shown in the attached video.
In addition, the trajectories under the different methods at $\Delta = -0.6$ are shown in Fig.\ref{fig:simulation environment}(b). 
From Fig.\ref{fig:ATE of simulation exeriments}, the RSDT's performance degraded when $\Delta$ decreased. 
Because RSDT does not consider the receding horizon optimization especially the motion smoothness cost, resulting in shaking rotations. 
The proportion of featureless areas in the image increased when $\Delta$ decreased, and this made the shaking rotation of planned camera view influence the localization accuracy more obviously.
Fig.\ref{fig:yaw_angle} shows that the yaw angle of the RSDT method changed with high-frequency ripples and brought sudden motions of the gimbal camera, degrading the feature matching and tracking of the SLAM.
Thanks to the consideration of degeneration and motion smoothness in the horizontal time window, the IGLOV planner performed much more smoothly, and the camera views were almost consistent with the motion direction because the yaw angles were smaller than 0.5 rad. 
Moreover, the views also turned to the feature-rich regions by the informative planning, this makes the camera by IGLOV locate well with the smallest estimation error, as shown in Fig.\ref{fig:ATE of simulation exeriments}.

\begin{figure}[htbp]
  \setlength{\abovecaptionskip}{0.cm} 
  \vspace{-0.2cm}  
	\centerline{\includegraphics[width=3.2in]{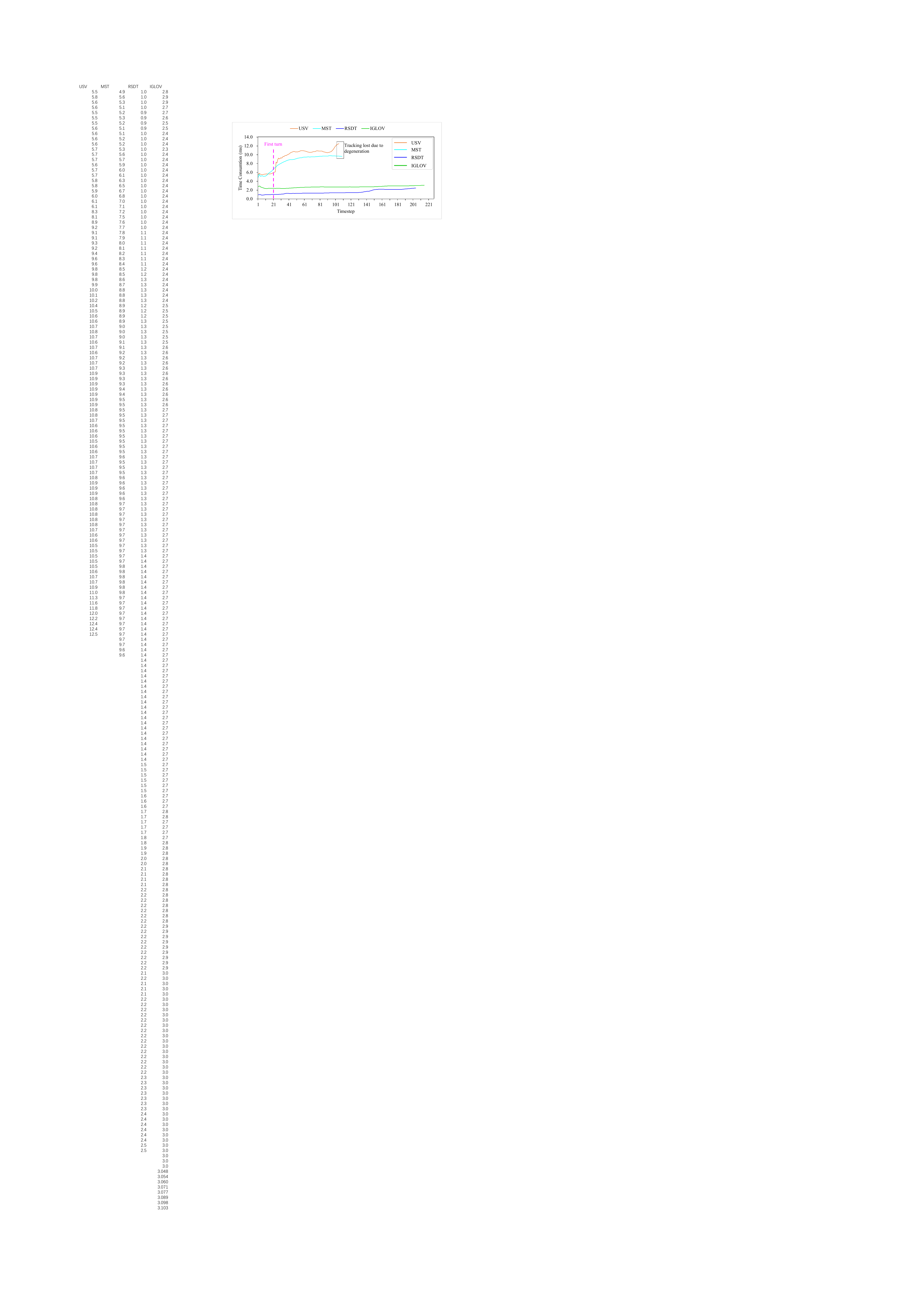}}
  \caption{Computation costs of different methods in one planning period.}
	\label{fig:time_consuming_of_different_methods}
  \vspace{-0.2cm}  
\end{figure}

The computation costs of different methods are compared in Fig.\ref{fig:time_consuming_of_different_methods}.
The time costs of the USV and MST methods positively correlate with the map size; the sudden increase of computation time occurred at the first turn of the trajectory because many new map points were added to the maps. 
{While the RSDT and IGLOV methods cost less time because the regular sampling method brings less sampling and evaluation. 
Besides, the IGLOV planner takes more time than the RSDT method due to the polynomial fitting and optimization.}

\begin{figure}[htbp]
  \setlength{\abovecaptionskip}{0.cm} 
  \vspace{-0.2cm}  
  \centering
  \subfigure[Simulation environment]{\includegraphics[height=2in]{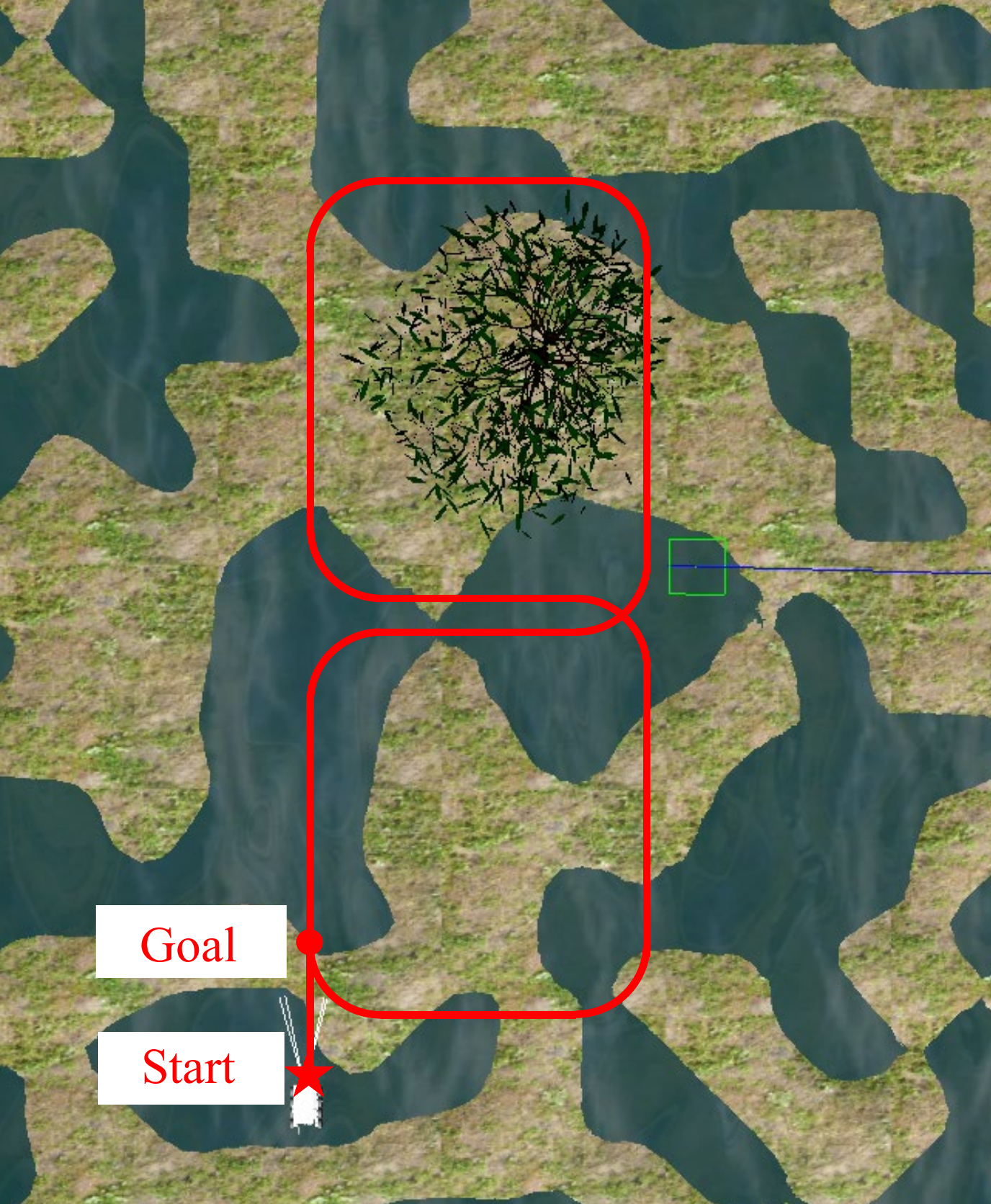}}
  \subfigure[Estimated trajectories]{\includegraphics[height=2in]{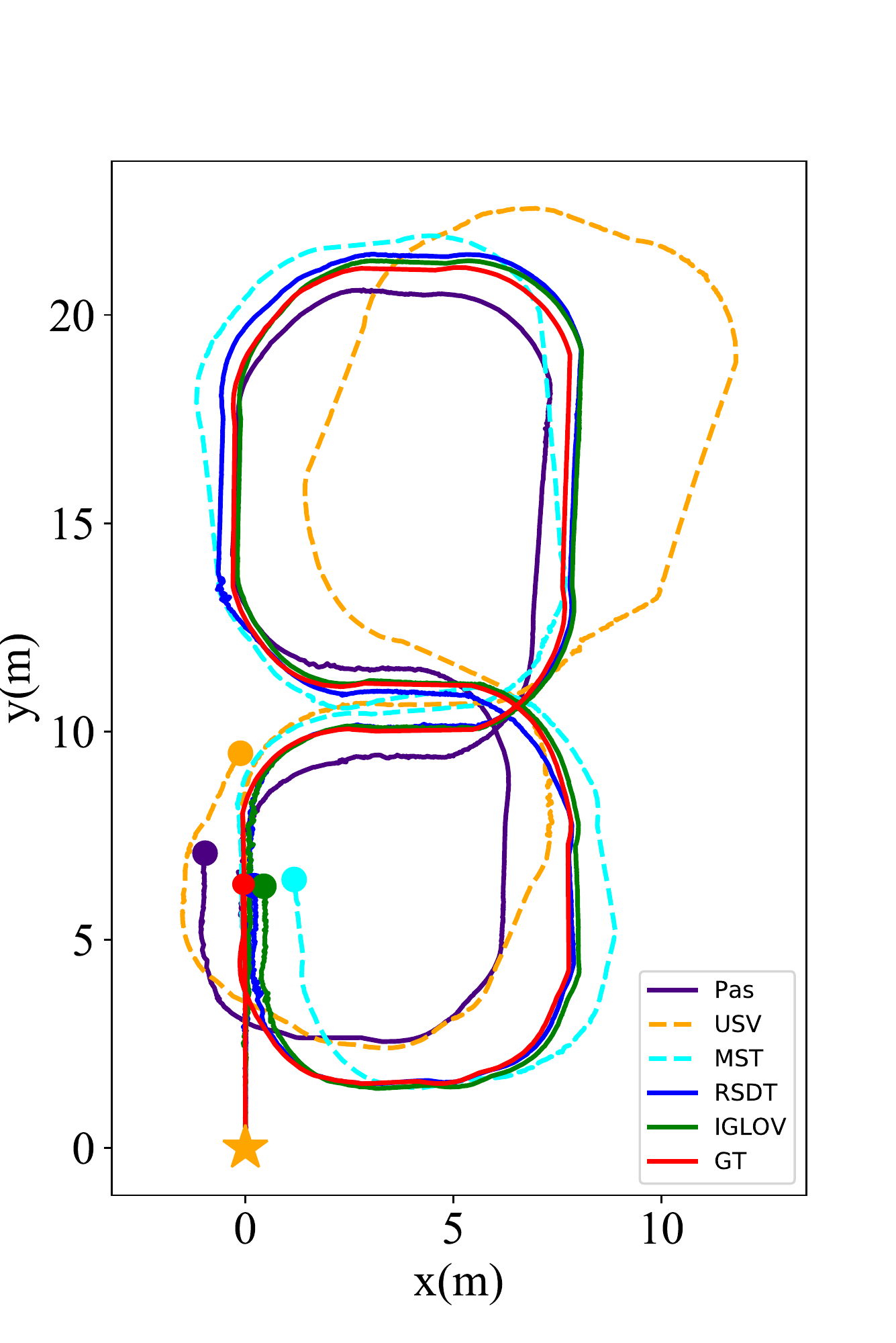}}
  \caption{ Simulation 2 and estimated trajectories under different methods. (a) Several featureless puddles lay in a rough grass plane, and a tree lays in the environment center. The robot was required to travel along the red trajectory. (b) Estimated trajectories under different methods. (c) Localization accuracy of different methods.  }
  \label{fig:simulation terrain environment}
  \vspace{-0.4cm}  
\end{figure}

\begin{table}[htbp]
  \setlength{\abovecaptionskip}{0.cm}
  \renewcommand\arraystretch{1.2}
  \caption{Localization accuracy of Simulation 2. }
  \centering
  \begin{tabular}{cccccc}
  \Xhline{1pt}
  \textbf{}      & Pas    & USV   & MST         & RSDT            & IGLOV           \\ \Xhline{1pt}
  {Mean(m)}         & 0.783     & 1.036     & 0.343    & 0.194    & \textbf{0.139}          \\ \Xhline{0.5pt}
  {RMSE(m)}         & 0.954     & 1.234     & 0.413    & 0.219    & \textbf{0.167} \\   \Xhline{1pt}
  \end{tabular}
  \label{table:error of sim2}
\end{table}

We further designed another simulation in a common wild environment, as shown in Fig.\ref{fig:simulation terrain environment}(a). The estimated trajectories, as well as the ground truth, are shown in Fig.\ref{fig:simulation terrain environment}(b). 
The PAS method exhibited large errors because the featureless water surface occupied the camera FOV in some parts of the trajectory. 
Although the view-landing-points under the MST and USV methods lay on the feature-rich regions, most of them lead the camera view opposed the motion direction; therefore, the two methods exhibited large errors due to the degeneration problem. 
In contrast, the RSDT and IGLOV methods performed well in this environment. 
TABLE \ref{table:error of sim2} shows that {the IGLOV planner has the minimal localization error than other methods because the estimation uncertainty, exploration and motion smoothness are considered.}

\vspace{-1em}
\subsection{Experiments}
Further, we designed several real-world experiments to verify the proposed approach with the experimental ground vehicle shown in Fig.\ref{fig:Robot Flatform}(b). 
Because both the USV and MST methods led to SLAM failures in our experiments,
we only presented the comparison between the passive method and the proposed one. 
We performed the first experiment in a typical outdoor campus environment with the robot moving along the campus road, while the second experiment was performed on a hillside with rough grass terrain.  

\subsubsection{Experiment 1}
\begin{figure}[htbp]
  \setlength{\abovecaptionskip}{0.cm} 
  \centering
  \includegraphics[width=3.45in]{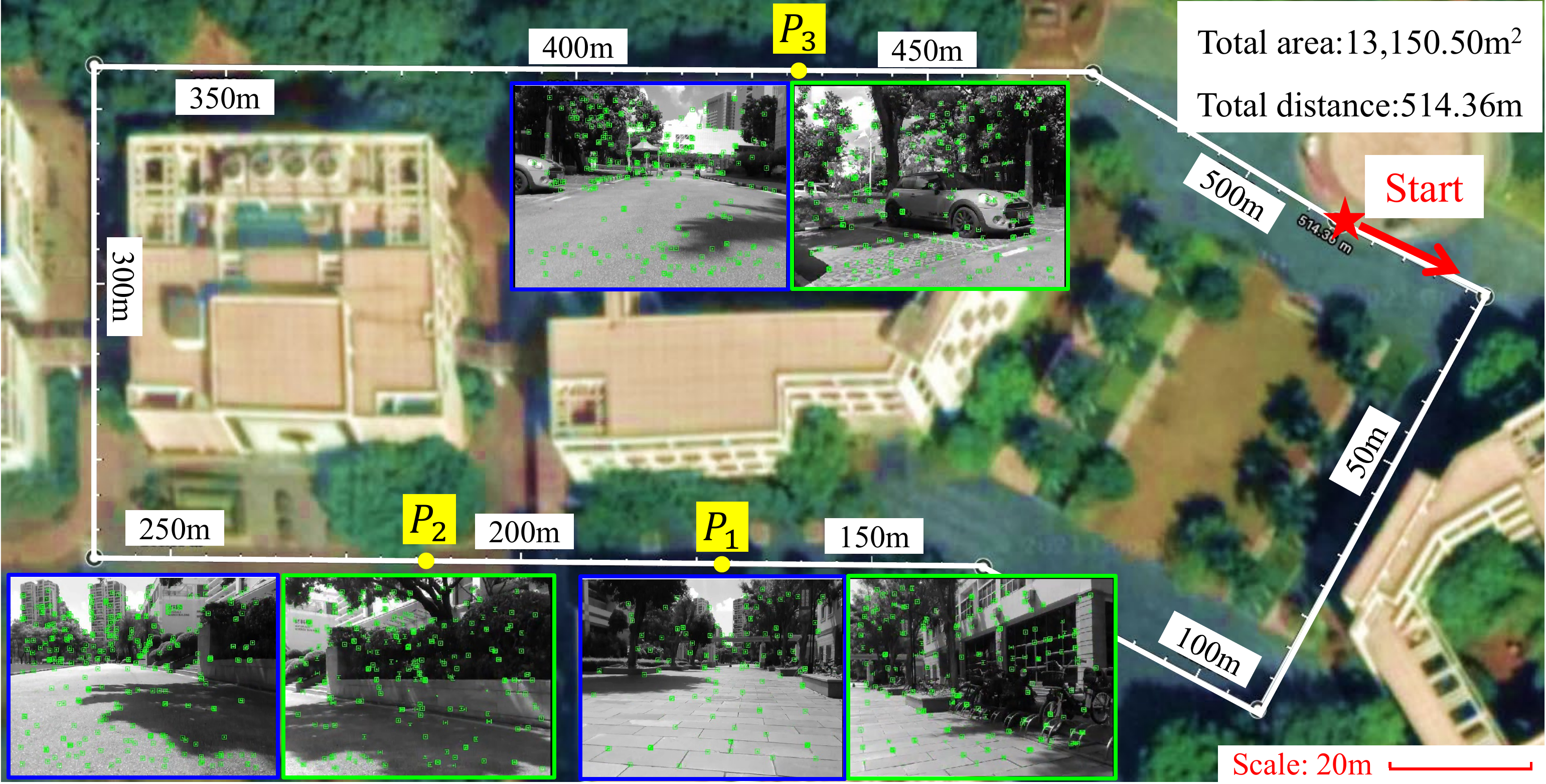}
  \caption{Bird's-eye view of the environment and the required trajectory in Experiment 1. The white line denotes the required trajectories. $P_1$-$P_3$ are the positions selected to show typical planning results. The subfigures with blue and green boxes are captured by the passive camera and the gimbal camera with IGLOV methods, respectively.  }
  \label{fig:Ex2-environment}
  \vspace{-0.6cm}  
\end{figure}

\begin{table}[htbp]
  \caption{Localization accuracy of Experiments. }
  \centering
  \begin{tabular}{ccccccc}
  \Xhline{1pt}
                      & \multicolumn{1}{l}{} & \multicolumn{1}{l}{Max(m)} & \multicolumn{1}{l}{Mean(m)} & \multicolumn{1}{l}{Min(m)} & \multicolumn{1}{l}{RMSE(m)} \\ \Xhline{1pt}
 \multirow{2}{*}{Ex1} & Pas                  & 2.552          & 0.892          & \textbf{0.041} & 1.027                       \\ \cline{2-6} 
                      & IGLOV                & \textbf{2.212} & \textbf{0.839} & 0.369          & \textbf{0.901}                        \\ \Xhline{1pt}
 \multirow{2}{*}{Ex2} & Pas                  & 2.289          & 1.242          & 0.341 & 1.322                        \\ \cline{2-6} 
                      & IGLOV                & \textbf{1.718} & \textbf{0.734} & \textbf{0.152}          & \textbf{0.814}                        \\ \Xhline{1pt}
 
  \end{tabular}
  \label{table:error of ex2}
  \vspace{-0.2cm}  
\end{table}

The first experiment (Ex1) was performed along a trajectory about 500m in Shenzhen University Town, as shown in Fig.\ref{fig:Ex2-environment}. 
The estimation error under the two methods are shown in TABLE \ref{table:error of ex2}. The IGLOV method performed better in mean and RMSE values. 
Compared to the simulation results in Section \ref{section simulation}.C, the localization accuracy of the passive method in the experiment was much closer to the IGLOV method. The reason is that the operating environment is urbanized and surrounded by feature-rich parterres, trees, and buildings; thus, the passive method also tracked well with sufficient features in each keyframe. 
However, when traveling along the trajectory, the IGLOV planner evaluated the information of the local environments and autonomously turned the camera view towards the areas with maximum information gain. 
Some typical planning results are demonstrated as the subfigures in Fig.\ref{fig:Ex2-environment}. 
The IGLOV method always turned the camera towards the local feature-rich areas, like the parked bicycles at position $P_1$, the high parterre at position $P_2$, and the parked car at position $P_3$. 
Although the improvement of localization accuracy was limited in the urbanized environments, the proposed method efficiently guided the camera view to local feature-rich areas.

\subsubsection{Experiment 2}

\begin{figure}[htbp]
  \setlength{\abovecaptionskip}{0.cm} 
  \centering
  {\includegraphics[width=3.0in]{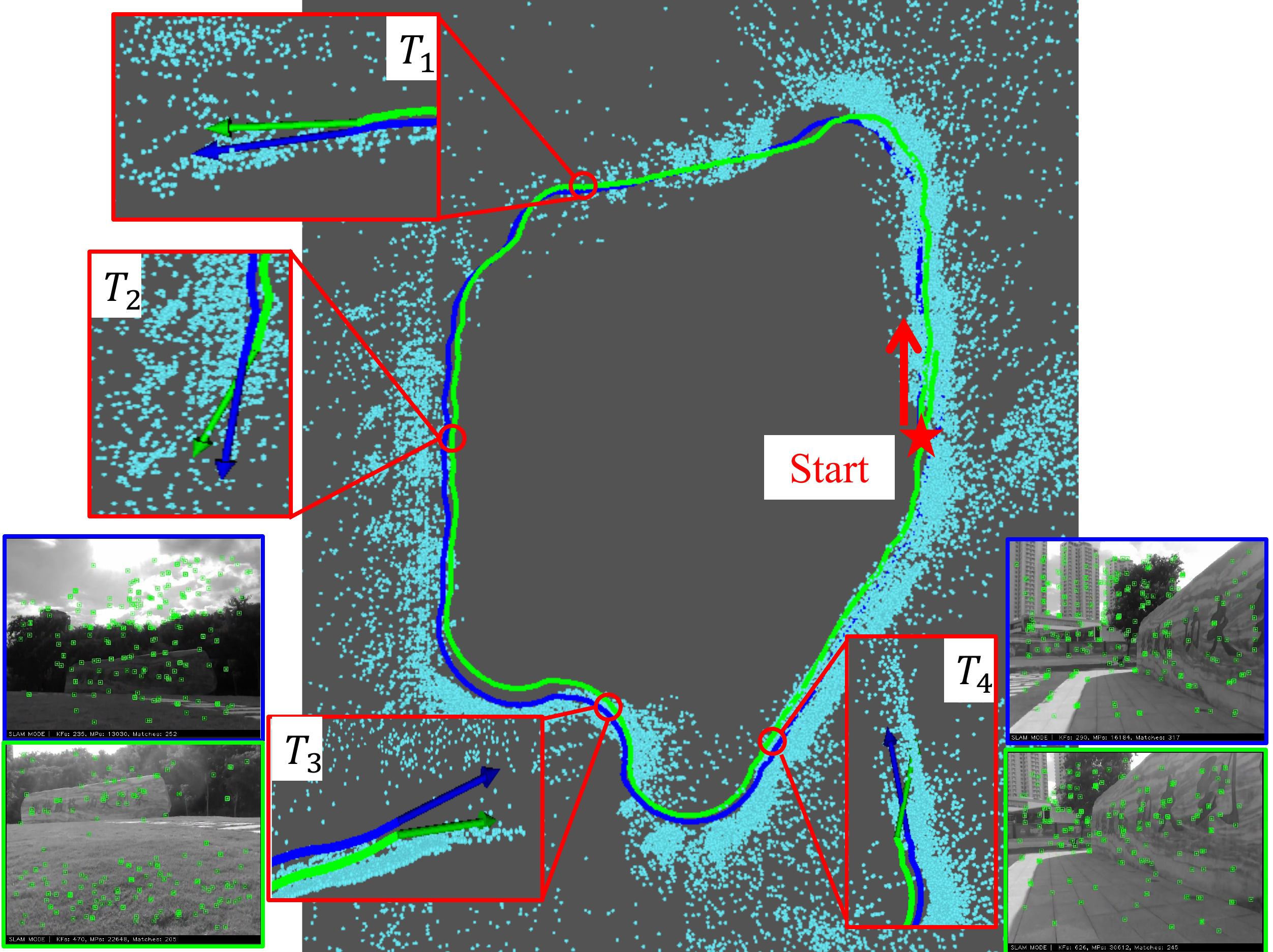}}
  \caption{Typical planning results of $T_1$-$T_4$ in Experiment 2. 
  The blue and green arrows in the subfigures denote the directions of the passive camera and the gimbal camera by IGLOV methods, respectively. The images with blue and green boxes are captured by the passive camera and the gimbal camera with IGLOV methods, respectively.  }
  \label{fig:Ex3 debug}
  \vspace{-0.4cm}  
\end{figure}

We further designed an experiment (Ex2) in a wild terrain environment to evaluate the performance of the proposed approach.
As shown in Fig.\ref{fig:Problem Formulation}, the experiment was performed on a hillside, the trajectory length was about 130m. 
The trajectory estimation results under the passive and the IGLOV methods, together with the ground truth, are shown in Fig.\ref{fig:Problem Formulation}. 
It is seen that the estimated trajectory under the IGLOV planner (green line) is closer to the ground truth (red line) than that under the passive method (blue line). The proposed approach is designed based on task space and feature points in the local sampling range; this makes the gimbal camera pay more attention to the feature-rich areas once it appears in the sampling range and then decreases the localization error.
TABLE \ref{table:error of ex2} shows the results for quantitative analysis. The visual SLAM was run without loop closure.
The IGLOV planner performed much better than the passive method; the reason is that the camera avoided the view to the featureless areas like the sky or the pavement and focused on local feature-rich areas like the steles or the trees.

The gimbal camera actively changed the view direction to the feature-rich regions shown as the subfigures in Fig.\ref{fig:Ex3 debug}. 
{Our approach stabilized the view direction towards the slope to complement the view direction swing in rough terrains at location $T_3$;
this avoided the glare caused by direct sunlight irradiation. In contrast, the sky occupied a large part of the camera FOV under the passive method.} With the help of the IGLOV method, the gimbal camera looked towards the feature-rich stele for helpful information at location $T_4$.

\section{Conclusion}\label{section6}
The paper develops a novel approach to realize active visual SLAM for ground or surface robots in challenging outdoor environments.
An information mapping algorithm is first proposed to represent the environmental information richness efficiently; the algorithm makes the online active view-planning possible. 
{A continuous information modeling method combined the regular sampling and polynomial regression is proposed to map the environmental information around the robot into multiple polynomial functions.
Based on the multiple polynomial functions, the informative planning problem is solved efficiently by numerical optimization.}
A receding horizon optimization-based method solves the view planning problem under degeneration and motion smoothness constraints.
Finally, several physical-engine simulations and experiments in outdoor environments are performed. The comparisons to the existing state-of-the-art approaches have verified the effectiveness of the proposed approach. 
{Our future work will focus on integrating robotic trajectory planning into the active SLAM for autonomous environmental exploration and informative navigation by considering map uncertainty, estimation of motion noise, infrared distribution and traversability simultaneously.} We will also study the robustness of vSLAM with the severe and high-frequency oscillations due to fast movement on rough terrains.

\addtolength{\textheight}{-0.2cm}   





\normalem
\bibliographystyle{IEEEtran}
\bibliography{IEEEabrv, bib/active_view_planning}

\begin{thebibliography}{10}
\providecommand{\url}[1]{#1}
\csname url@samestyle\endcsname
\providecommand{\newblock}{\relax}
\providecommand{\bibinfo}[2]{#2}
\providecommand{\BIBentrySTDinterwordspacing}{\spaceskip=0pt\relax}
\providecommand{\BIBentryALTinterwordstretchfactor}{4}
\providecommand{\BIBentryALTinterwordspacing}{\spaceskip=\fontdimen2\font plus
\BIBentryALTinterwordstretchfactor\fontdimen3\font minus
  \fontdimen4\font\relax}
\providecommand{\BIBforeignlanguage}[2]{{%
\expandafter\ifx\csname l@#1\endcsname\relax
\typeout{** WARNING: IEEEtran.bst: No hyphenation pattern has been}%
\typeout{** loaded for the language `#1'. Using the pattern for}%
\typeout{** the default language instead.}%
\else
\language=\csname l@#1\endcsname
\fi
#2}}
\providecommand{\BIBdecl}{\relax}
\BIBdecl

\bibitem{9690581}
R.~Duan, D.~P. Paudel, C.~Fu, and P.~Lu, ``Stereo orientation prior for uav
  robust and accurate visual odometry,'' \emph{IEEE/ASME Transactions on
  Mechatronics}, vol.~27, no.~5, pp. 3440--3450, 2022.

\bibitem{9830851}
J.~Liu, X.~Li, Y.~Liu, and H.~Chen, ``Rgb-d inertial odometry for a
  resource-restricted robot in dynamic environments,'' \emph{IEEE Robotics and
  Automation Letters}, vol.~7, no.~4, pp. 9573--9580, 2022.

\bibitem{murORB2}
R.~Mur-Artal and J.~D. Tard\'os, ``{ORB-SLAM2}: an open-source {SLAM} system
  for monocular, stereo and {RGB-D} cameras,'' \emph{IEEE Transactions on
  Robotics}, vol.~33, no.~5, pp. 1255--1262, 2017.

\bibitem{qin2018vins}
T.~Qin, P.~Li, and S.~Shen, ``Vins-mono: A robust and versatile monocular
  visual-inertial state estimator,'' \emph{IEEE Transactions on Robotics},
  vol.~34, no.~4, pp. 1004--1020, 2018.

\bibitem{wang2019robust}
Z.~Wang, J.~Zhang, S.~Chen, C.~Yuan, J.~Zhang, and J.~Zhang, ``Robust high
  accuracy visual-inertial-laser slam system,'' in \emph{2019 IEEE/RSJ
  International Conference on Intelligent Robots and Systems (IROS)}.\hskip 1em
  plus 0.5em minus 0.4em\relax IEEE, 2019, pp. 6636--6641.

\bibitem{matsuki2018omnidirectional}
H.~Matsuki, L.~Von~Stumberg, V.~Usenko, J.~St{\"u}ckler, and D.~Cremers,
  ``Omnidirectional dso: Direct sparse odometry with fisheye cameras,''
  \emph{IEEE Robotics and Automation Letters}, vol.~3, no.~4, pp. 3693--3700,
  2018.

\bibitem{zhang2016benefit}
Z.~Zhang, H.~Rebecq, C.~Forster, and D.~Scaramuzza, ``Benefit of large
  field-of-view cameras for visual odometry,'' in \emph{2016 IEEE International
  Conference on Robotics and Automation (ICRA)}.\hskip 1em plus 0.5em minus
  0.4em\relax IEEE, 2016, pp. 801--808.

\bibitem{frintrop2008attentional}
S.~Frintrop and P.~Jensfelt, ``Attentional landmarks and active gaze control
  for visual slam,'' \emph{IEEE Transactions on Robotics}, vol.~24, no.~5, pp.
  1054--1065, 2008.

\bibitem{chen2020active}
Y.~Chen, S.~Huang, and R.~Fitch, ``Active slam for mobile robots with area
  coverage and obstacle avoidance,'' \emph{IEEE/ASME Transactions on
  Mechatronics}, vol.~25, no.~3, pp. 1182--1192, 2020.

\bibitem{wang2022automated}
Y.~Wang, H.~Chen, S.~Zhang, and W.~Lu, ``Automated camera-exposure control for
  robust localization in varying illumination environments,'' \emph{Autonomous
  Robots}, vol.~46, no.~4, pp. 515--534, 2022.

\bibitem{deng2018feature}
X.~Deng, Z.~Zhang, A.~Sintov, J.~Huang, and T.~Bretl, ``Feature-constrained
  active visual slam for mobile robot navigation,'' in \emph{2018 IEEE
  International Conference on Robotics and Automation (ICRA)}.\hskip 1em plus
  0.5em minus 0.4em\relax IEEE, 2018, pp. 7233--7238.

\bibitem{khosoussi2019reliable}
K.~Khosoussi, M.~Giamou, G.~S. Sukhatme, S.~Huang, G.~Dissanayake, and J.~P.
  How, ``Reliable graphs for slam,'' \emph{The International Journal of
  Robotics Research}, vol.~38, no. 2-3, pp. 260--298, 2019.

\bibitem{2020arXiv200803324Z}
Z.~{Zhang} and D.~{Scaramuzza}, ``{Fisher Information Field: an Efficient and
  Differentiable Map for Perception-aware Planning},'' \emph{arXiv e-prints},
  p. arXiv:2008.03324, Aug. 2020.

\bibitem{hornung2013octomap}
A.~Hornung, K.~M. Wurm, M.~Bennewitz, C.~Stachniss, and W.~Burgard, ``Octomap:
  An efficient probabilistic 3d mapping framework based on octrees,''
  \emph{Autonomous robots}, vol.~34, no.~3, pp. 189--206, 2013.

\bibitem{jadidi2018gaussian}
M.~G. Jadidi, J.~V. Miro, and G.~Dissanayake, ``Gaussian processes autonomous
  mapping and exploration for range-sensing mobile robots,'' \emph{Autonomous
  Robots}, vol.~42, no.~2, pp. 273--290, 2018.

\bibitem{zhang2018perception}
Z.~Zhang and D.~Scaramuzza, ``Perception-aware receding horizon navigation for
  mavs,'' in \emph{2018 IEEE International Conference on Robotics and
  Automation (ICRA)}.\hskip 1em plus 0.5em minus 0.4em\relax IEEE, 2018, pp.
  2534--2541.

\bibitem{strader2020perception}
J.~Strader, K.~Otsu, and A.-a. Agha-mohammadi, ``Perception-aware autonomous
  mast motion planning for planetary exploration rovers,'' \emph{Journal of
  Field Robotics}, vol.~37, no.~5, pp. 812--829, 2020.

\bibitem{kim2015active}
A.~Kim and R.~M. Eustice, ``Active visual slam for robotic area coverage:
  Theory and experiment,'' \emph{The International Journal of Robotics
  Research}, vol.~34, no. 4-5, pp. 457--475, 2015.

\bibitem{zhou2021fuel}
B.~Zhou, Y.~Zhang, X.~Chen, and S.~Shen, ``Fuel: Fast uav exploration using
  incremental frontier structure and hierarchical planning,'' \emph{IEEE
  Robotics and Automation Letters}, vol.~6, no.~2, pp. 779--786, 2021.

\bibitem{dharmadhikari2020motion}
M.~Dharmadhikari, T.~Dang, L.~Solanka, J.~Loje, H.~Nguyen, N.~Khedekar, and
  K.~Alexis, ``Motion primitives-based path planning for fast and agile
  exploration using aerial robots,'' in \emph{2020 IEEE International
  Conference on Robotics and Automation (ICRA)}.\hskip 1em plus 0.5em minus
  0.4em\relax IEEE, 2020, pp. 179--185.

\bibitem{zhu2021online}
H.~Zhu, J.~J. Chung, N.~R. Lawrance, R.~Siegwart, and J.~Alonso-Mora, ``Online
  informative path planning for active information gathering of a 3d surface,''
  in \emph{2021 IEEE International Conference on Robotics and Automation
  (ICRA)}.\hskip 1em plus 0.5em minus 0.4em\relax IEEE, 2021, pp. 1488--1494.

\bibitem{indelman2015planning}
V.~Indelman, L.~Carlone, and F.~Dellaert, ``Planning in the continuous domain:
  A generalized belief space approach for autonomous navigation in unknown
  environments,'' \emph{The International Journal of Robotics Research},
  vol.~34, no.~7, pp. 849--882, 2015.

\bibitem{barfoot2017state}
T.~D. Barfoot, \emph{State estimation for robotics}.\hskip 1em plus 0.5em minus
  0.4em\relax Cambridge University Press, 2017.

\bibitem{chen2021anchor}
Y.~Chen, L.~Zhao, Y.~Zhang, S.~Huang, and G.~Dissanayake, ``Anchor selection
  for slam based on graph topology and submodular optimization,'' \emph{IEEE
  Transactions on Robotics}, vol.~38, no.~1, pp. 329--350, 2021.

\bibitem{zhang2019beyond}
Z.~Zhang and D.~Scaramuzza, ``Beyond point clouds: Fisher information field for
  active visual localization,'' in \emph{2019 International Conference on
  Robotics and Automation (ICRA)}.\hskip 1em plus 0.5em minus 0.4em\relax IEEE,
  2019, pp. 5986--5992.

\bibitem{bishop2010optimality}
A.~N. Bishop, B.~Fidan, B.~D. Anderson, K.~Do{\u{g}}an{\c{c}}ay, and P.~N.
  Pathirana, ``Optimality analysis of sensor-target localization geometries,''
  \emph{Automatica}, vol.~46, no.~3, pp. 479--492, 2010.

\bibitem{pukelsheim2006optimal}
F.~Pukelsheim, \emph{Optimal design of experiments}.\hskip 1em plus 0.5em minus
  0.4em\relax SIAM, 2006.

\bibitem{boyd2004convex}
S.~Boyd, S.~P. Boyd, and L.~Vandenberghe, \emph{Convex optimization}.\hskip 1em
  plus 0.5em minus 0.4em\relax Cambridge university press, 2004.

\bibitem{proenca2018probabilistic}
P.~F. Proenca and Y.~Gao, ``Probabilistic rgb-d odometry based on points, lines
  and planes under depth uncertainty,'' \emph{Robotics and Autonomous Systems},
  vol. 104, pp. 25--39, 2018.

\bibitem{grupp2017evo}
M.~Grupp, ``evo: Python package for the evaluation of odometry and slam,''
  \emph{url: https://github. com/MichaelGrupp/evo}, 2017.

\bibitem{zeng2020pc}
R.~Zeng, W.~Zhao, and Y.-J. Liu, ``Pc-nbv: A point cloud based deep network for
  efficient next best view planning,'' in \emph{2020 IEEE/RSJ International
  Conference on Intelligent Robots and Systems (IROS)}.\hskip 1em plus 0.5em
  minus 0.4em\relax IEEE, 2020, pp. 7050--7057.

\bibitem{palomeras2018autonomous}
N.~Palomeras, N.~Hurt{\'o}s, M.~Carreras, and P.~Ridao, ``Autonomous mapping of
  underwater 3-d structures: From view planning to execution,'' \emph{IEEE
  Robotics and Automation Letters}, vol.~3, no.~3, pp. 1965--1971, 2018.

\end{thebibliography}

\vspace{-43pt}
\begin{IEEEbiography}[{\includegraphics[width=1in,height=1.25in,clip,keepaspectratio]{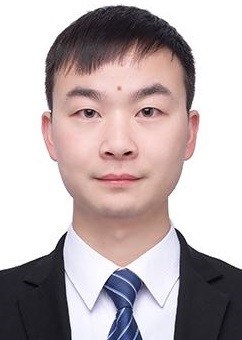}}]{Zhihao Wang}
  received the B.Eng. degree in mechanical engineering from the WuHan University of Technology, WuHan, China, in 2017, and the M.Eng. degree in control engineering from the Huazhong University of Science and Technology, WuHan, China, in 2019. He is currently working toward the Ph.D. degree in robotics with the School of Mechatronics Engineering and Automation, Harbin Institute of Technology, Shenzhen, China.

  His research interests lie in robotic motion planning and control, 3D exploration and reconstruction.
\end{IEEEbiography}

\vspace{-43pt}
\begin{IEEEbiography}[{\includegraphics[width=1in,height=1.25in,clip,keepaspectratio]{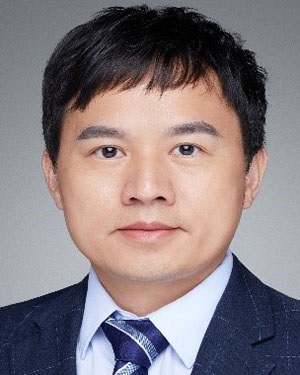}}]{Haoyao Chen}
  received the B.Eng. degree in mechatronics and automation from the University of Science and Technology of China, Hefei, China, in 2004, and the Ph.D. degree in robotics and automation from the University of Science and Technology of China and the City University of Hong Kong, Hong Kong, in 2009. 
  
  He is currently a Professor with the Harbin Institute of Technology Shenzhen, Shenzhen, China, and the State Key Laboratory of Robotics and System, Harbin, China. His research interests lie in visual servoing, multirobot systems, motion control, and aerial manipulation.
\end{IEEEbiography}

\vspace{-43pt}
\begin{IEEEbiography}[{\includegraphics[width=1in,height=1.25in,clip,keepaspectratio]{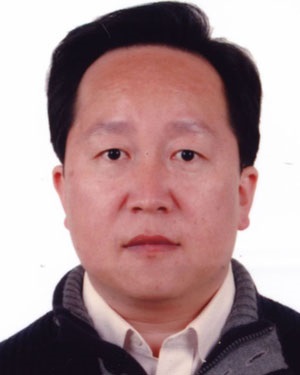}}]{Shiwu Zhang}
  (Member, IEEE) received the B.S. degree in mechanical and electrical engineering and the Ph.D. degree in precision instrumentation and precision machinery from the University of Science and Technology of China (USTC), Hefei, China, in 1997 and 2003, respectively. 
  
  He is currently a Professor with the Department of Precision Machinery and Precision Instrumentation, USTC. He is the author of more than 100 papers in different journals and conferences. His research interests include smart materials and their applications in bio-inspired robots, amphibious robot, soft robots, and terradynamics.
\end{IEEEbiography}

\vspace{-43pt}
\begin{IEEEbiography}[{\includegraphics[width=1in,height=1.25in,clip,keepaspectratio]{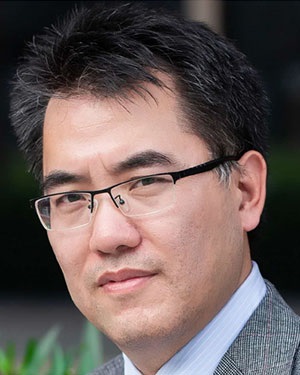}}]{Yunjiang Lou}
  (Senior Member, IEEE) received the B.S. and M.E. degrees in automation from the University of Science and Technology of China, Hefei, China, in 1997 and 2000, respectively, and the Ph.D. degree in electrical and electronic engineering from the Hong Kong University of Science and Technology, Hong Kong, in 2006. 
  
  He is currently with the State Key Laboratory of Robotics and Systems and the Shenzhen Key Laboratory for Advanced Motion Control and Modern Automation Equipments, School of Mechatronics Engineering and Automation, Harbin Institute of Technology (Shenzhen), Shenzhen, China. His research interests include motion control, mechanism design, compliant actuators, and industrial robots.
\end{IEEEbiography}

\end{document}